\documentclass[lettersize,journal]{IEEEtran}
\usepackage{amsmath,amsfonts}
\usepackage{algorithmic}
\usepackage{algorithm}
\usepackage{array}
\usepackage{textcomp}
\usepackage{stfloats}
\usepackage{url}
\usepackage{verbatim}
\usepackage{graphicx}
\usepackage{cite}
\usepackage{amsmath}
\usepackage{svg}
\usepackage{graphicx}
\usepackage{multirow}
\usepackage[most]{tcolorbox}
\usepackage{tabularx}
\usepackage{tabularray}
\usepackage{lineno,hyperref}
\usepackage{threeparttable}
\usepackage{xcolor}
\usepackage{booktabs}
\usepackage{amsfonts} 
\usepackage{amsmath}
\usepackage{caption}
\usepackage{subcaption}
\usepackage{enumitem}
\usepackage{tablefootnote}
\usepackage{placeins}

\usepackage{booktabs,arydshln}
\makeatletter
\def\adl@drawiv#1#2#3{%
        \hskip.5\tabcolsep
        \xleaders#3{#2.5\@tempdimb #1{1}#2.5\@tempdimb}%
                #2\z@ plus1fil minus1fil\relax
        \hskip.5\tabcolsep}
\newcommand{\cdashlinelr}[1]{%
  \noalign{\vskip\aboverulesep
           \global\let\@dashdrawstore\adl@draw
           \global\let\adl@draw\adl@drawiv}
  \cdashline{#1}
  \noalign{\global\let\adl@draw\@dashdrawstore
           \vskip\belowrulesep}}
\makeatother

\usepackage{leftidx}

\usepackage{pifont}
\newcommand{\cmark}{\ding{51}}%
\newcommand{\xmark}{\ding{55}}%

\begin{document}
\title{Continuous Urban Change Detection from Satellite Image Time Series with Temporal Feature Refinement and Multi-Task Integration}

\author{Sebastian Hafner, Heng Fang, Hossein Azizpour and Yifang Ban*, \textit{Senior Member}, \textit{IEEE}\thanks{The authors are with the Division of Geoinformatics (Sebastian Hafner and Yifang Ban) and the Division of Robotics, Perception and Learning (Heng Fang and Hossein Azizpour) at KTH Royal Institute of Technology, 114 28 Stockholm, Sweden.}\thanks{*Corresponding author: Yifang Ban (e-mail: yifang@kth.se)}}

\markboth{}%
{Shell \MakeLowercase{\textit{et al.}}: A Sample Article Using IEEEtran.cls for IEEE Journals}


\maketitle

\begin{abstract}
Urbanization advances at unprecedented rates, leading to negative environmental and societal impacts. Remote sensing can help mitigate these effects by supporting sustainable development strategies with accurate information on urban growth. Deep learning-based methods have achieved promising urban change detection results from optical satellite image pairs using convolutional neural networks (ConvNets), transformers, and a multi-task learning setup. However, bi-temporal methods are limited for continuous urban change detection, i.e., the detection of changes in consecutive image pairs of satellite image time series (SITS), as they fail to fully exploit multi-temporal data ($>$ 2 images). Existing multi-temporal change detection methods, on the other hand, collapse the temporal dimension, restricting their ability to capture continuous urban changes. Additionally, multi-task learning methods lack integration approaches that combine change and segmentation outputs. To address these challenges, we propose a continuous urban change detection framework incorporating two key modules. The temporal feature refinement (TFR) module employs self-attention to improve ConvNet-based multi-temporal building representations. The temporal dimension is preserved in the TFR module, enabling the detection of continuous changes. The multi-task integration (MTI) module utilizes Markov networks to find an optimal building map time series based on segmentation and dense change outputs. The proposed framework effectively identifies urban changes based on high-resolution SITS acquired by the PlanetScope constellation (F1 score 0.551), Gaofen-2 (F1 score 0.440), and WorldView-2 (F1 score 0.543). Moreover, our experiments on three challenging datasets demonstrate the effectiveness of the proposed framework compared to bi-temporal and multi-temporal urban change detection and segmentation methods. Code is available on GitHub: \url{https://github.com/SebastianHafner/ContUrbanCD}.
\end{abstract}


\IEEEraisesectionheading{\section{Introduction}\label{sec:introduction}}

\IEEEPARstart{U}{rbanization} is progressing at unprecedented rates \cite{liu2020high}. Thus, the global amount of urban land is projected to increase by a factor of 2--6 over the 21st century \cite{gao2020mapping}. The rapid expansion of urban land, i.e., urban sprawl, is associated with multiple negative effects on the environment and human well-being \cite{johnson2001environmental, sarkodie2020global}. To mitigate urban sprawl, informed and sustainable urban development strategies are crucial \cite{arshad2022quantifying}. However, these strategies are currently hampered by a lack of timely information on the extent of urban land. 

Remote sensing is an efficient tool to monitor the Earth's surface \cite{ban2016change}. Urban changes are commonly detected from two satellite images acquired at different times over the same geographical area. Traditional change detection methods use arithmetic operations to derive change features from bi-temporal image pairs. For example, various arithmetic methods have been developed to derive change features from optical images, such as image differencing, image regression, and change vector analysis \cite{lu2004change}. These features are then classified into changed/unchanged pixels or objects using different classification algorithms, including machine learning algorithms \cite{lu2004change, ban2016change}.

\begin{figure}[!h]
     \centering
     \begin{subfigure}[b]{0.35\textwidth}
         \centering
         \includegraphics[width=\textwidth]{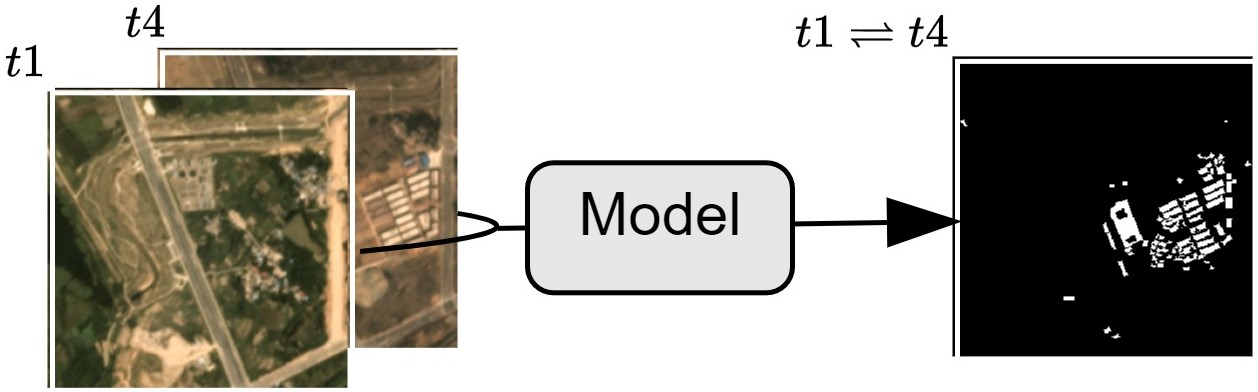}
         \caption{Bi-temporal urban change detection}
         \label{fig:bitemporal_cd}
     \end{subfigure}
     \vspace{0.5cm}
     \vfill
     \begin{subfigure}[b]{0.4\textwidth}
         \centering
         \includegraphics[width=\textwidth]{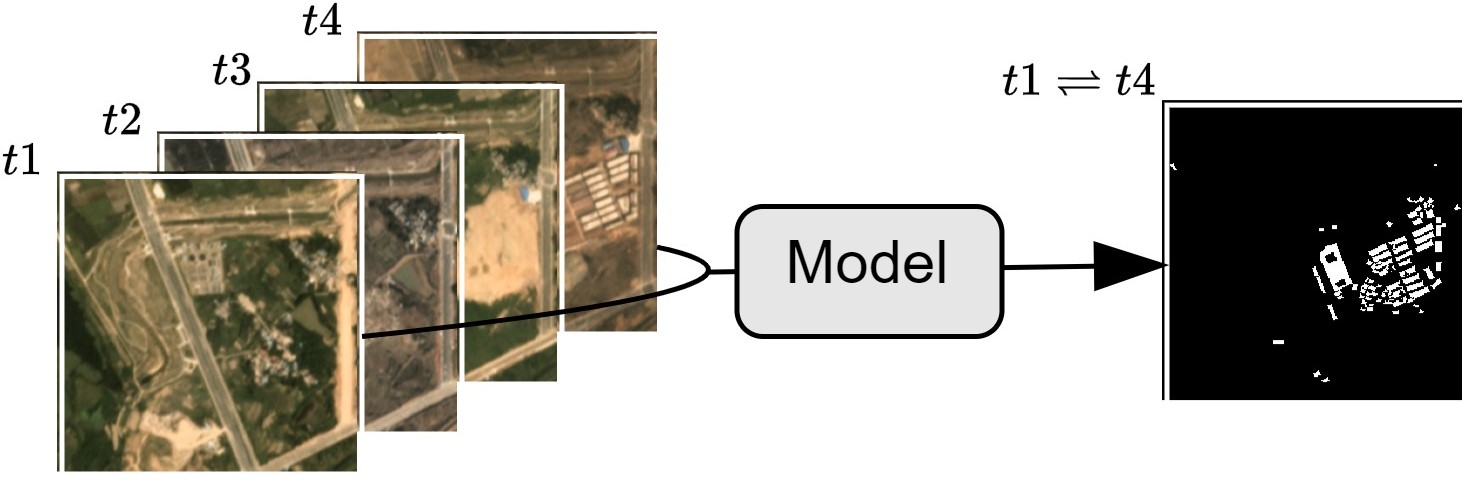}
         \caption{Multi-temporal urban change detection}
         \label{fig:multitemporal_cd}
     \end{subfigure}
     \vspace{0.5cm}
    \vfill
     \begin{subfigure}[b]{0.45\textwidth}
         \centering
         \includegraphics[width=\textwidth]{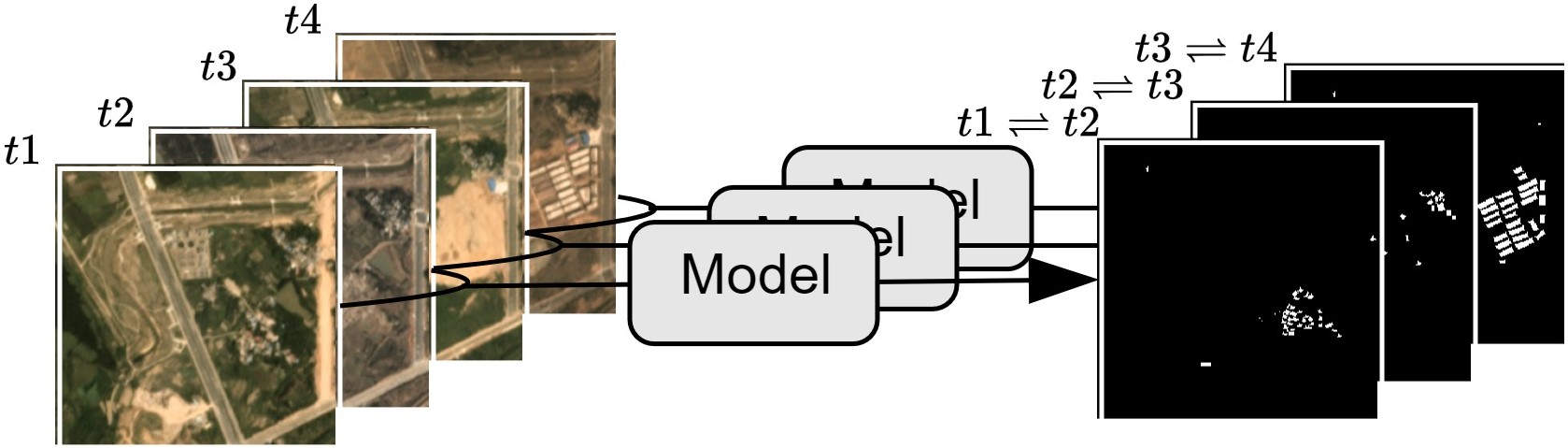}
         \caption{Continuous urban change detection using a bi-temporal model}
         \label{fig:cucd_bitemporal}
     \end{subfigure}
     \vspace{0.5cm}
    \vfill
     \begin{subfigure}[b]{0.45\textwidth}
         \centering
         \includegraphics[width=\textwidth]{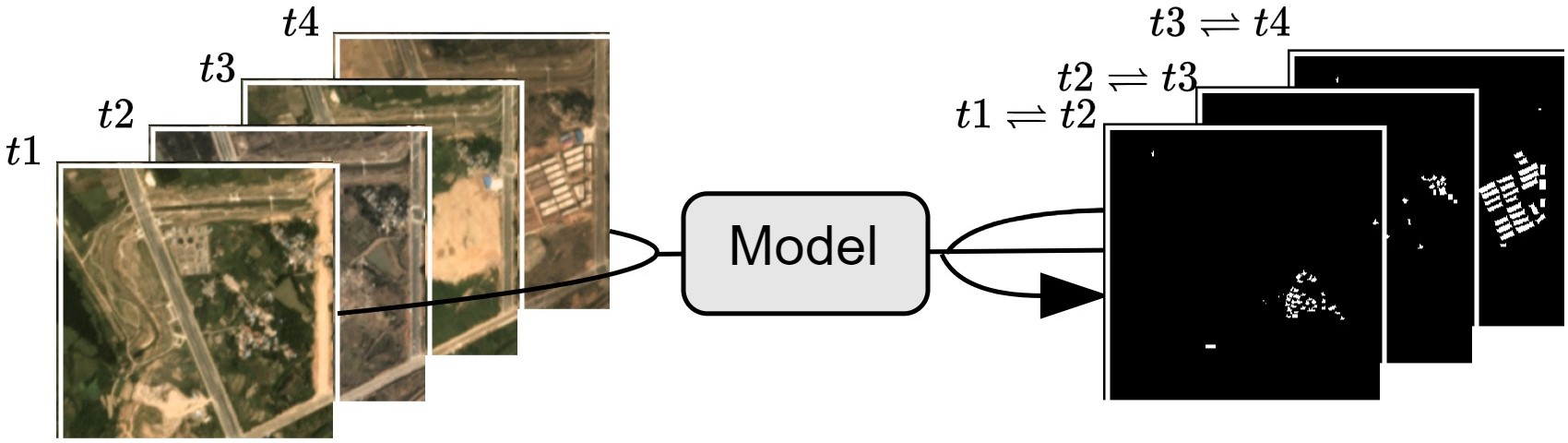}
         \caption{Continuous urban change detection using the proposed model}
         \label{fig:cucd_proposed}
     \end{subfigure}
     \hfill
    \caption{Overview of standard urban change detection frameworks and the proposed method. (a) Bi-temporal urban change detection is typically performed on image pairs acquired multiple years apart. (b) Multi-temporal change detection methods leverage image time series but only predict changes between the first and last image. (c) A bi-temporal model is applied to consecutive image pairs of a time series to perform continuous urban change detection; however, this method fails to incorporate multi-temporal (i.e., $>2$ images) information (c). On the other hand, we propose (d) a continuous urban change detection method that incorporates multi-temporal information.}
    \label{fig:cd_methods}
\end{figure}

In recent years, deep learning has been continuously replacing traditional change detection methods \cite{zhu2017deep, jiang2022survey, lv2024novel, lv2024spatial}. Specifically, deep convolutional neural networks (ConvNets) have been used extensively for change detection in bi-temporal optical satellite image pairs (see Fig. \ref{fig:bitemporal_cd}). The simplest way of adapting common ConvNets such as U-Net \cite{ronneberger2015u} for change detection is with an input-level fusion (or early fusion \cite{daudt2018urban}) strategy, referring to the concatenation of image pairs before passing them to a ConvNet. Contrarily, late fusion strategies typically process images separately in a Siamese network consisting of two ConvNets with shared weights. Extracted bi-temporal features are then fused using concatenation or absolute differencing \cite{daudt2018urban, daudt2018fully}. Since Siamese networks are generally considered preferable to input-level fusion strategies, multiple studies developed modules that are incorporated into Siamese network architectures to improve feature representations \cite{fang2021snunet, basavaraju2022ucdnet, liu2023attention}. For example, Chen \textit{et al.} \cite{chen2020spatial, chen2021remote} proposed to refine features extracted by ConvNets from very high-resolution (VHR) imagery using a transformer-based module, alleviating the limited long-range context modeling capability of convolutions with self-attention. Since then, self-attention has become a popular mechanism for capturing long-range spatial dependencies in VHR change detection \cite{bandara2022transformer, liu2023attention, marsocci2023inferring, noman2024remote}.

In recent years, high-resolution (i.e., 1 -- 10 m) satellite image time series (SITS) have become increasingly available \cite{roy2021global}. Those data have enabled a shift from detecting land cover changes in image pairs acquired years apart to continuous annual and sub-annual change detection \cite{zhu2022remote}. In contrast, urban change detection methods are predominantly designed for bi-temporal change detection from image pairs acquired multiple years apart \cite{shafique2022deep}. However, considering the unprecedented rate of global urbanization \cite{liu2018high}, it is essential to develop a new suite of methods that detect urban changes continuously. While continuous change detection can be achieved by applying a bi-temporal model to consecutive image pairs in SITS (see Fig. \ref{fig:cucd_bitemporal}), this approach fails to exploit multi-temporal, i.e., $>$ 2 images, information. Furthermore, recent studies \cite{papadomanolaki2019detecting, papadomanolaki2021deep} have demonstrated the effectiveness of multi-temporal change detection models that predict changes between the first and last images of a SITS (see Fig. \ref{fig:multitemporal_cd}). For example, multi-temporal information can help to reduce commission errors from registration errors, illumination differences, or other types of change unrelated to the problem of interest \cite{papadomanolaki2021deep}. In addition, it can mitigate the effect of cloud artifacts in single images \cite{papadomanolaki2019detecting}. Existing multi-temporal change detection methods employ either recurrent layers, such as long short-term memory (LSTM) layers \cite{hochreiter1997long}, or 3D convolutional layers to model temporal information \cite{papadomanolaki2019detecting, papadomanolaki2021deep, meshkini2024multi, saha2020change}. While these layers effectively model short-range temporal dependencies in time series data, the self-attention mechanism can explicitly model temporal dependencies across all timestamps of a time series \cite{vaswani2017attention}. Thus, several recent segmentation methods for SITS employ the self-attention mechanism to explicitly model temporal dependencies across all timestamps of a time series \cite{garnot2021panoptic, tarasiou2023vits}. However, the temporal models in these methods collapse the temporal dimension, resulting in a single output feature. Therefore, they do not facilitate continuous urban change detection, which requires the full temporal information to produce change maps between each consecutive image pair in the SITS.

Another promising avenue of research for change detection is multi-task learning \cite{caruana1997multitask}, where a related semantic segmentation task is trained parallel to the change detection task using a shared feature representation. The change detection task is typically combined with building segmentation for urban change detection \cite{liu2020building, papadomanolaki2021deep, hafner2022urban, shu2022mtcnet, hafner2023semi}. To that end, Siamese networks are extended with an additional decoder for the semantic segmentation task. The feature maps extracted by the encoder are then shared between the change decoder and the segmentation decoder. However, despite the attention multi-task learning has attracted in change detection, effective methods to integrate segmentation and change outputs have been largely unaddressed. For example, most multi-task urban change detection studies consider building and change predictions independent outputs of the network \cite{liu2020building, papadomanolaki2021deep, hafner2022urban, shu2022mtcnet, hafner2023semi}. Therefore, these studies do not account for inconsistencies between the building segmentation and urban change predictions. Moreover, they fail to exploit the complementary information produced by multi-task predictions.

In this paper, we propose a continuous urban change detection method (see Fig. \ref{fig:cucd_proposed}) and explore two research gaps in the current literature, namely (1) the modeling of multi-temporal information using self-attention for continuous urban change detection and (2) the integration of segmentation and change predictions in multi-task learning setups. Specifically, we propose a new network architecture that relies on convolutions to extract multi-temporal building representations and employs self-attention to model temporal dependencies in feature space while preserving the temporal dimension. We also propose a novel integration approach that determines the optimal building segmentation for each image in a time series based on the multi-task network outputs. The effectiveness of the proposed architecture and integration approach is demonstrated on three urban change detection datasets featuring high-resolution optical SITS. 

The following summarizes the main contributions of this paper.

\begin{itemize}
    \item We introduce a continuous urban change detection model that produces change outputs for each consecutive image pair in SITS while leveraging the full temporal dimension of the time series.
    \item To enable continuous change detection, we present a transformer-based feature refinement module that effectively models temporal information in SITS. Importantly, our module preserves the temporal dimension of the representations, in contrast to existing temporal modules that collapse multi-temporal representations into a single one.
    \item We propose a new multi-task integration approach that represents segmentation and change outputs in Markov networks to find the optimal building maps time series based on the network outputs. 
    \item Experiments on three datasets, namely SpaceNet 7, WUSU, and TSCD, show that the proposed continuous urban change detection method is more effective than related methods.
\end{itemize}

\section{Related Work}\label{sec:related}
\def\pwid{.2\linewidth}

\subsection{Bi-temporal change detection}

In recent years, a plethora of deep learning-based bi-temporal change detection methods have been proposed. Most of these works focus on developing new Siamese network architectures and/or training strategies. Initially, Daudt \textit{et al.} \cite{daudt2018urban, daudt2018fully} proposed two Siamese ConvNet architectures for change detection. The Siam-Diff and Siam-Conc architectures employ encoders with shared weights for feature extraction from bi-temporal high-resolution image pairs and combine the corresponding feature maps using a subtraction and concatenation strategy, respectively. While encoders and decoders in these models follow the U-Net architecture \cite{ronneberger2015u}, Fang \textit{et al.} \cite{fang2021snunet} incorporated a nested U-Net (i.e., UNet++ \cite{zhou2018unet++}) into a Siamese network to maintain high-resolution, fine-grained representations through dense skip connections. Many works also improved Siamese networks by incorporating different modules into the architecture. For example, an ensemble channel attention module was proposed for feature refinement in \cite{fang2021snunet}, and a new spatial pyramid pooling block was utilized in \cite{basavaraju2022ucdnet} to preserve shapes of change areas. 

However, most recent methods are developed for bi-temporal change detection from VHR image pairs. Consequently, many methods employ the self-attention mechanism to improve the modeling of long-range dependencies in VHR imagery \cite{chen2020spatial, chen2021remote, liu2023attention, bandara2022transformer}. Both \cite{chen2020spatial} and \cite{chen2021remote} extract image features with ConvNets and employ self-attention modules to learn more discriminative features. Other works combined ConvNets and transformers with attention modules and multi-scale processing \cite{liu2023attention, wang2023iftsdnet}. Bandara \textit{et al.} \cite{bandara2022transformer}, on the other hand, proposed a fully transformer-based change detection method. Specifically, ChangeFormer combines two hierarchically structured transformer encoders with shared weights and a multi-layer perception decoder in a Siamese network architecture. Since transformer-based methods strongly rely on pretraining, Noman \textit{et al.} \cite{noman2024remote} recently proposed ScratchFormer which is a transformer-based change detection method that is trained from scratch but achieves SOTA performance. The ScrachFormer architecture utilizes shuffled sparse attention layers that enable faster convergence due to their sparse structure. Although these transformer-based methods are considered SOTA for urban change detection, their effectiveness has been predominately demonstrated on bi-temporal VHR datasets such as LEVIR-CD \cite{chen2020spatial} and WHU-CD \cite{ji2018fully}. In comparison, high-resolution imagery is acquired much more frequently by satellite constellations such as PlanetScope, making it an invaluable data source for change detection applications. Therefore, developing methods that effectively leverage transformers for change detection from high-resolution imagery is crucial.

\subsection{Change detection and segmentation from time series data}

Few studies have developed deep learning methods for urban change detection from high-resolution SITS. For example, Papadomanolaki \textit{et al.} \cite{papadomanolaki2019detecting} proposed to incorporate LSTM networks into a U-Net model to leverage optical SITS for change detection. Their L-UNet outperformed bi-temporal ConvNet-based methods on a bi-temporal dataset enriched with intermediate satellite images \cite{papadomanolaki2019detecting}. Others proposed an encoder-decoder LSTM model that is trained to rearrange temporally shuffled time series \cite{saha2020change}. The core assumption of this unsupervised method is that the model fails to correctly rearrange shuffled data for changed pixels. On the other hand, Meshkini \textit{et al.} \cite{meshkini2024multi} proposed a weakly supervised change detection method that employs 3D convolutional layers to capture spatial-temporal information in SITS. Recently, He \textit{et al.} \cite{he2024time} 
presented a deep learning method for time series land cover change detection. However, since their model only uses one-dimensional convolutions along the temporal dimension, it does not consider the spatial dimension, which is a limiting factor for high-resolution data. 

Due to the limited number of change detection methods for SITS, we also expand this review to the semantic segmentation of SITS. Several recent semantic segmentation methods for SITS employed the self-attention mechanism for temporal modeling of multi-temporal features \cite{garnot2021panoptic, tarasiou2023vits, cai2024cost}. Garnot \textit{et al.} \cite{garnot2021panoptic} employed a lightweight-temporal attention encoder \cite{garnot2020lightweight} for the temporal modeling of multi-temporal feature maps extracted using a shared ConvNet encoder. Similarly, Cai \textit{et al.} \cite{cai2024cost} employed an attention bidirectional LSTM module for temporal modeling of ConvNet-based feature maps time series. The modules in both studies collapse the temporal dimension, resulting in a single feature map obtained using a ConvNet decoder. While Tarasiou \textit{et al.} \cite{tarasiou2023vits} used a vision transformer to learn feature representation from SITS, their model also outputs a single feature map for semantic segmentation.

In summary, existing multi-temporal change detection methods rely on recurrent and 3D convolutional layers for temporal modeling. While multi-temporal semantic segmentation methods frequently employ temporal attention-based modules, they collapse the temporal dimension similarly to multi-temporal change detection methods. While these methods can be adapted for change detection, collapsing the temporal dimension limits them to the detection of changes between the first and the last image of a SITS (i.e., multi-temporal change detection in Fig. \ref{fig:multitemporal_cd}).

\subsection{Multi-task learning}

Multi-task learning has been investigated by several studies for urban change detection over the past years. Liu \textit{et al.} \cite{liu2020building} proposed a dual-task Siamese ConvNet to learn more discriminative feature representations for building change detection from bi-temporal image pairs. The proposed dual-task constrained deep Siamese convolutional network (DTCDSCN) consists of three main components: a shared ResNet-based encoder, a shared decoder for building segmentation, and a separate decoder for change detection. On the other hand, Papadomanolaki \textit{et al.} \cite{papadomanolaki2021deep} proposed a multi-task learning framework for urban change detection from image time series by adding building segmentation tasks for the first and last images of a time series to the urban change detection task. While L-UNet \cite{papadomanolaki2019detecting} is employed to extract changes, the segmentation is performed with a separate decoder that directly uses the feature maps extracted for the image pair by the shared encoder. 

Some urban change detection studies also combined multi-task learning with semi-supervised learning \cite{hafner2022urban, shu2022mtcnet}. In \cite{hafner2022urban}, the Siam-Diff network \cite{daudt2018fully} was extended with an additional shared decoder for building segmentation, and an unsupervised term was introduced to encourage consistency between the changes derived from the building predictions and those predicted by the change decoder. Shu \textit{et al.} \cite{shu2022mtcnet}, on the other hand, proposed to learn consistency between two building predictions corresponding to the pre-change image. The first prediction is obtained by segmenting the pre-change image and the second one by combining segmentation features of the post-change image with changes features. 

In general, these multi-task studies demonstrate that learning a segmentation task in parallel to the change detection task improves the latter. However, none of these studies investigate combining the change and segmentation network outputs to improve performance. Consequently, inconsistencies between the network outputs are also not accounted for.

\section{The proposed method}
\label{sec:method}

\subsection{Overview}

\begin{figure*}[!h]
\centering
\includegraphics[width=\textwidth]{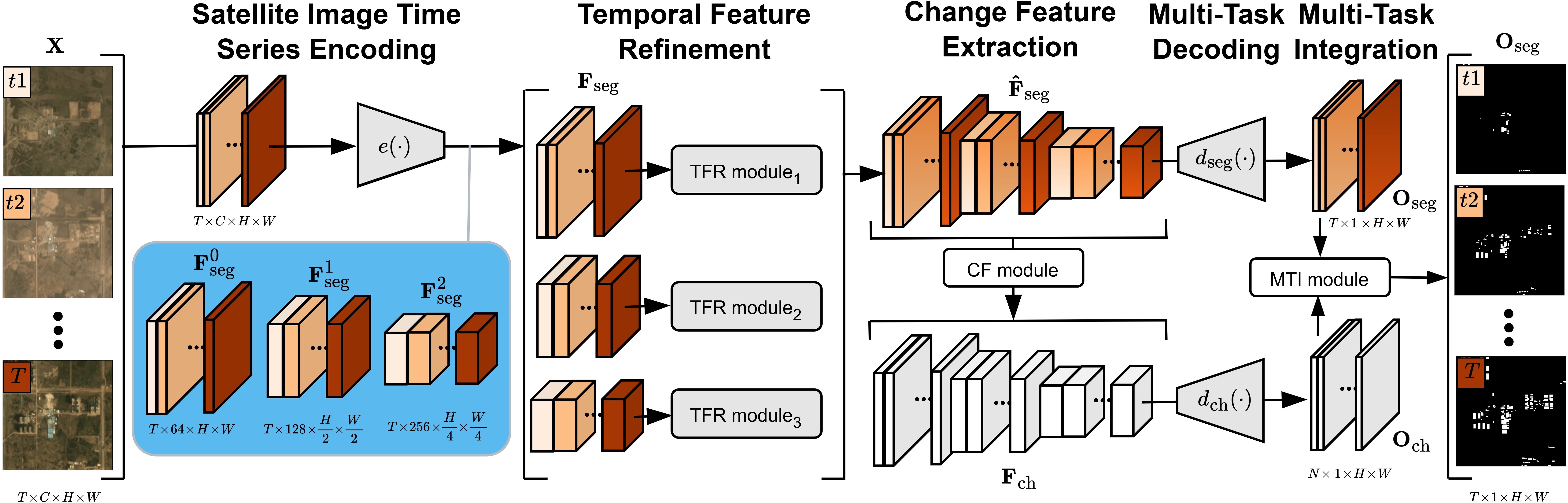}
\caption{Overview of the proposed method. First, an encoder extracts multi-scale feature maps from a satellite image time series. Next, transformer-based temporal feature refinement (TFR) modules enrich feature maps at each scale with multi-temporal information, and the change feature (CF) module generates bi-temporal difference feature maps from the temporally refined feature maps. Then, two separate decoders are used to obtain segmentation and change predictions from the respective feature maps. Finally, predictions for the two tasks are combined using a multi-task integration (MTI) module. For brevity, only three out of the five scales of the feature maps are shown.}
\label{fig:overview}
\end{figure*}

As illustrated in Fig. \ref{fig:overview}, the main components of the proposed method are a ConvNet-based encoder, transformer-based temporal feature refinement (TFR) modules, a change feature (CF) module, two task-specific ConvNet-based decoders, and a Markovian multi-task integration (MTI) module. The following summarizes the urbanization monitoring process of the proposed method for a time series of satellite images:

\begin{enumerate}[label=\arabic*.,ref=\arabic*]
    \item First, for each image in the time series, multi-scale feature maps are extracted using an encoder with shared weights.
    
    \item Next, the above time series of feature maps are grouped by scale and fed to separate TFR modules consisting of transformer encoder layers. The temporally refined feature maps are regrouped according to their timestamp.

    \item Then, the CF module obtains change feature maps from the temporally refined segmentation feature maps. The module considers changes between all possible combinations of temporal pairs. 
    
    \item Two task-specific decoders are deployed to obtain building segmentation outputs for each image in the time series from the temporally refined segmentation features maps and change outputs from the change feature maps.

    \item Finally, the building and urban change outputs are combined using the MTI module. The module uses pixel-wise Markov networks to obtain optimal building states for the SITS.
\end{enumerate}

Detailed descriptions of the components comprising the proposed method, as well as the loss function, are given in the following sections.

\subsection{Satellite image time series encoding}

We consider a time series of $T$ satellite images, represented as $\mathbf{X} \in \mathbb{R}^{T \times C \times H \times W}$, where $C$, $H$, and $W$ denote the channel, height, and width dimensions, respectively. A ConvNet encoder with shared weights is utilized to separately extract feature maps $\mathbf{F}_{\rm seg}$ from each image in the time series, as follows:

\begin{equation}
    \mathbf{F}_{\rm seg} = e(\mathbf{X}),
\end{equation}

\noindent where $e(\cdot)$ represents the encoder, and subscript ${\rm seg}$ indicates that the feature maps contain representations for building segmentation.

The architecture of the encoder is based on the U-Net encoder \cite{ronneberger2015u}. Specifically, after an initial convolution block, the combination of a max-pooling layer and a consecutive convolution block is applied four times. Each of these four steps halves the spatial dimensions $H$ and $W$ due to the pooling operation, whereas the number of features $D$ is doubled with the convolution block. Importantly, U-Net achieves precise localization by leveraging skip connections that forward the feature maps before each pooling operation to the decoder. Therefore, the output of the encoder consists of five feature map time series with different scales. To denote the scale of these time series, we introduce superscript $s$ in the notation: $\mathbf{F}_{\rm seg}^{s}$, where $s \in \{0, 1, 2, 3, 4\}$. For a given feature map time series $\mathbf{F}_{\rm seg}^{s}$, the sizes of the height and width dimensions, as well as the feature dimension, are dependent on $s$, as follows:

\begin{equation}
    H^s = \frac{H}{2^{s}}, \thinspace W^s = \frac{W}{2^{s}}, \thinspace D^s = 64 \cdot 2^s.
\end{equation}

It should be noted that for brevity, Fig. \ref{fig:overview} illustrates the proposed method for $s \in \{0, 1, 2\}$.


\subsection{Temporal feature refinement}

\begin{figure}[!h]
\centering
\includegraphics[width=.45\textwidth]{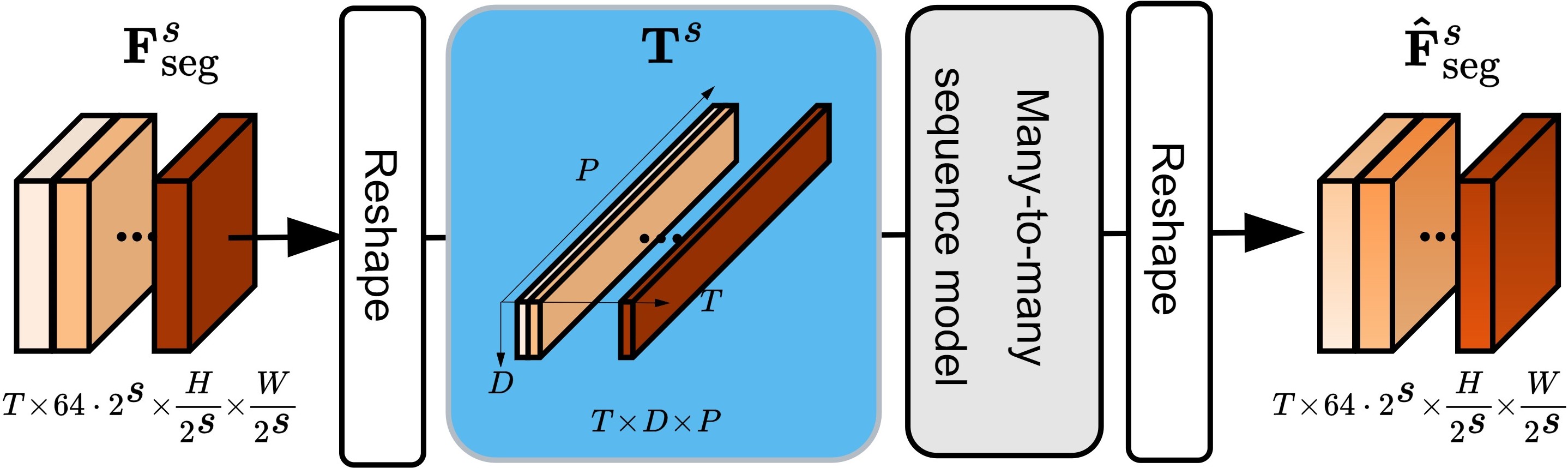}
\caption{Illustration of the temporal feature refinement (TFR) module, preserving the temporal dimension of the input feature maps.}
\label{fig:trf_module}
\end{figure}

The TFR module, illustrated in Fig. \ref{fig:trf_module}, creates temporally refined feature maps using the self-attention mechanism along the temporal dimension \cite{vaswani2017attention}. Unlike the temporal modules in existing change detection and segmentation methods for SITS, our module preserves the temporal dimension.

The module takes as input a time series of feature maps at the same scale $s$ and reshapes this 4D tensor to a 3D tensor $\mathbf{T}^s \in \mathbb{R}^{T \times D^s \times P^s}$ by flattening the spatial dimensions $H^s$ and $W^s$. Consequently, $T$, $D$, and $P$ represent the temporal, feature embedding, and spatial dimensions, respectively. 
After reshaping the feature map time series, self-attention is applied along the temporal dimension $T$ for each cell in the spatial dimension $P$. However, since the self-attention mechanism contains no recurrence, it is necessary to first inject information about the temporal position of the feature vectors in the time series. Specifically, temporal encodings having the same dimension as the feature vectors are generated based on sine and cosine functions of different frequencies \cite{vaswani2017attention}. These relative temporal encodings are then added to the feature vectors.

\begin{figure}[!h]
\centering
\includegraphics[width=.45\textwidth]{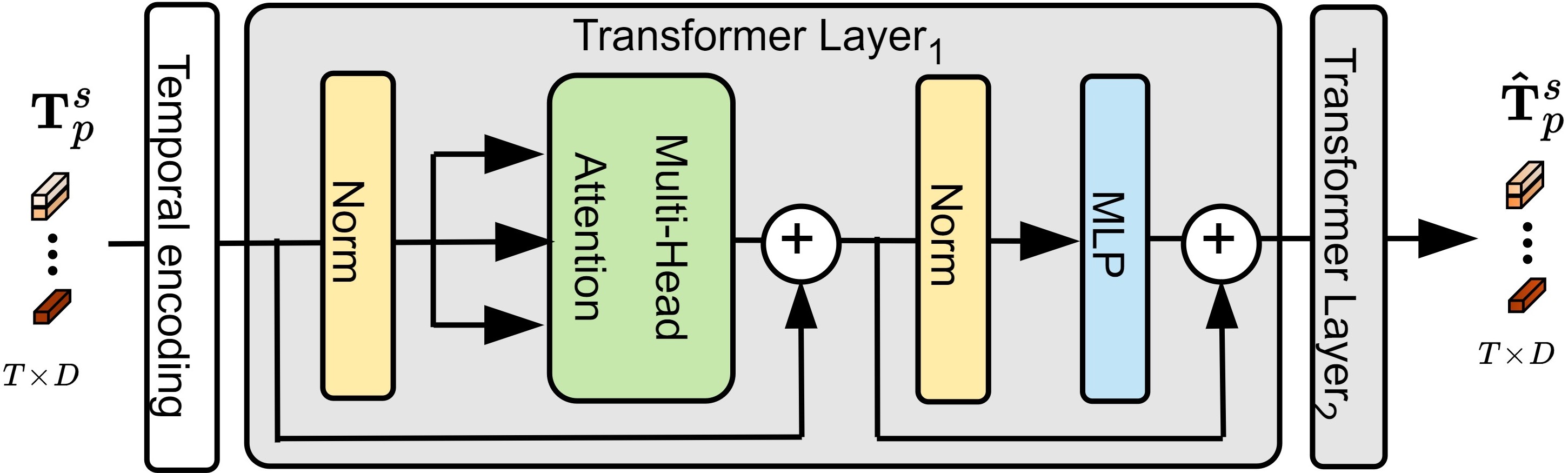}
\caption{Illustration of the transformer encoder layer applying multi-headed self-attention.}
\label{fig:transformer_encoder}
\end{figure}

The tensor, enriched with relative temporal position information, is passed through two transformer encoder layers (see Fig. \ref{fig:transformer_encoder}). The key component of the transformer encoder layer is the multi-head attention block, which performs self-attention defined as follows:

\begin{equation}
\label{eq:attention}
    \text{Att}(\mathbf{Q}, \mathbf{K}, \mathbf{V}) = \text{softmax}\left(\frac{\mathbf{Q} \mathbf{K}^{T}}{\sqrt{D}} \right)\mathbf{V},
\end{equation}

\noindent where $\mathbf{Q}$, $\mathbf{K}$, and $\mathbf{V}$ are referred to as query, key, and value, respectively. The query-key-value triplet is computed with three linear projection layers with parameter matrices $\mathbf{W}^Q, \mathbf{W}^K, \mathbf{W}^V \in \mathbb{R}^{D \times D}$ that are separately applied to a given cell of the 3D tensor $\mathbf{T}^s_p$, where $p$ denotes the cell index in $\mathbf{T}^s$.

The core idea of multi-head attention is, however, that self-attention is performed multiple times in parallel using $h$ attention heads, as follows:

\begin{equation}
\begin{split}
    \text{MultiHead}(\mathbf{Q}, \mathbf{K}, \mathbf{V}) & = \text{Concat}\left(\text{head}_1, ..., \text{head}_h\right)\mathbf{W}^O, \\
    \text{where } \text{head}_i & = \text{Att}(\mathbf{Q}\mathbf{W}_i^Q, \mathbf{K}\mathbf{W}_i^K, \mathbf{V}\mathbf{W}_i^V).
\end{split}
\end{equation}

Each $\text{head}_i$ performs self-attention on different projections of the keys, values, and queries obtained from linear layers with parameter matrices $\mathbf{W}_i^Q, \mathbf{W}_i^K, \mathbf{W}_i^V \in \mathbb{R}^{D \times D_{\rm head}}$. Finally, the concatenated outputs of the heads are reprojected using parameter matrix $\mathbf{W}^Q \in \mathbb{R}^{hD_{\rm head} \times D}$. We employ $h = 2$ attention heads, where the head dimension is given by $D_{\rm head} = D / h$.

After applying self-attention to each multi-temporal feature vector, we obtain a 3D tensor containing temporally refined building representations. In practice, however, all cells of tensor $\mathbf{T}^s$ are processed in parallel by incorporating them into the batch dimension which is omitted for clarity. Finally, the 3D tensor is reshaped to the dimensions of the feature map time series by unflattering dimension $P$. We denote this temporally refined feature map time series with $\mathbf{\hat{F}}^{s}_{\rm seg}$.


\subsection{Change feature extraction}

The CF module is used to convert the temporally refined segmentation feature maps to change feature maps. Specifically, we consider the temporally refined feature maps $\mathbf{\hat{F}}_{\rm seg}^t$ and $\mathbf{\hat{F}}_{\rm seg}^k$, corresponding to two images acquired at time $t$ and time $k$, where $1 \leq t < k \leq T$. Then, change feature map $\mathbf{F}_{\rm ch}^n$ corresponding to the urban changes between the bi-temporal image pair is constructed as follows:

\begin{equation}
    \mathbf{F}_{\rm ch}^n = \mathbf{\hat{F}}_{\rm seg}^k - \mathbf{\hat{F}}_{\rm seg}^t,
\end{equation}

\noindent where $n$ denotes a change edge between timestamps $t$ and $k$. It should be noted that this is done for each scale $s$ of the feature maps.

This operation is identical to the change feature computation in the Siam-Diff method \cite{daudt2018fully}. However, instead of only considering changes between the first and the last images of a time series, the CF module computes change feature maps for all possible combinations of image pairs. The total number of combinations $N_{\rm dense}$ is given by the length of the time series $T$, defined as follows:

\begin{equation}
\label{eq:comb}
\begin{aligned}
N_{\rm dense} = &\frac{T(T - 1)}{2}.  \\
\end{aligned}
\end{equation}

We refer to the scenario when all possible combinations of image pairs are considered as \textit{dense}. However, we also investigate sparser settings such as \textit{adjacent}, considering only changes between temporally adjacent images ($N_{ \rm adjacent} = T - 1$), and \textit{cyclic}, adding the changes between the first and last image to the adjacent setting ($N_{\rm cyclic} = T$) (see Fig. \ref{fig:mrf_adjacent} -- \ref{fig:mrf_dense} for visualizations of these settings in Markov networks with $T = 5$). We provide ablation results on the different edge settings in Tables \ref{tab:ablation_loss} and \ref{tab:ablation_mti}. Finally, it should be noted that the CF module does not have any trainable parameters.

\subsection{Multi-task decoding}

Two separate decoders are deployed to convert the temporally enriched segmentation feature maps and the change feature maps to building outputs and urban change outputs, respectively. Formally, we obtain $T$ built-up area segmentation outputs $\mathbf{O}_{\rm seg} \in (0, 1)^{T \times H \times W}$ with the segmentation decoder $d_{\rm seg}(\cdot)$ as follows:

\begin{equation}
    \mathbf{O}_{\rm seg} = d_{\rm seg}(\mathbf{\hat{F}}_{\rm seg}).
\end{equation}

Furthermore, we obtain $N$ change detection outputs $\mathbf{O}_{\rm ch} \in (0, 1)^{N \times H \times W}$ with the change decoder $d_{\rm ch}(\cdot)$ as follows:

\begin{equation}
    \mathbf{O}_{\rm ch} = d_{\rm ch}(\mathbf{F}_{\rm ch}).
\end{equation}

Both decoders follow the architecture of the U-Net expansive path consisting of four upsampling blocks followed by a \(1 \times 1\) convolution layer and a sigmoid activation function. Upsampling blocks double the height and width of feature maps via a transpose conv \(2 \times 2\) layer. Upsampled feature maps are then concatenated along the channel dimension with the temporally refined feature maps matching their scale (skip connection). Subsequently, the layer triplet \(3 \times 3\) convolution, batch normalization, and ReLu activation is applied twice.

\subsection{Loss function}
\label{subsec:loss_function}

The network is trained using a loss function composed of two terms, namely for the urban change detection task ($\mathcal{L}_{\rm ch}$) and the building segmentation task ($\mathcal{L}_{\rm seg}$). For both loss terms, a Jaccard metric measuring the similarity between continuous network outputs $\mathbf{O} \in (0, 1)$ and binary labels $\mathbf{Y} \in \{0, 1\}$ is used \cite{duque2021power}. We denote the Jaccard metric by $J(,)$ and define the loss function as follows:

\begin{equation}
\label{eq:sup_loss}
\mathcal{L} = \sum_{t=1}^{T} J(\mathbf{O}^t_{\rm seg}, \mathbf{Y}^t_{\rm seg}) + \sum_{n=1}^{N} J(\mathbf{O}_{\rm ch}^{n}, \mathbf{Y}_{\rm ch}^{n}), \\
\end{equation}

\noindent where $T$ denotes the length of the time series and $N$ denotes the number of edges (i.e., combinations of bi-temporal image pairs) considered. Segmentation and change labels are denoted by $\mathbf{Y}_{\rm seg} \in \{0, 1\}^{T \times H \times W}$ and $\mathbf{Y}_{\rm ch} \in \{0, 1\}^{N \times H \times W}$, respectively. Specifically, we assume that pixel-wise building annotations, $\mathbf{Y}_{\rm seg}$, are available and derive pixel-wise built-up changes, $\mathbf{Y}_{\rm ch}$, according to the considered edges.

\subsection{Multi-task integration}

To combine segmentation and change predictions, we propose the MTI module which determines the optimal building segmentation output for each image in a time series. Since this is a pixel-based approach, we represent the location of a specific pixel in the segmentation and change output notations by introducing superscript coordinates $i$ and $j$. Following that, $\mathbf{O}_{\rm seg}^{(i, j), t}$ denotes the segmentation output for a specific pixel $i, j$ at timestamp $t$, where $i \in \{1,\ldots,H\}$, $j \in \{1,\ldots,W\}$, and $t \in \{1,\ldots,T\}$. Likewise, the change output for a specific pixel is denoted by  $\mathbf{O}_{\rm ch}^{(i, j),n}$, where $n$ denotes the change edge connecting timestamps $t$ and $k$.

\begin{figure}[!h]
     \centering
     \begin{subfigure}[b]{0.45\textwidth}
         \centering
         \includegraphics[width=\textwidth]{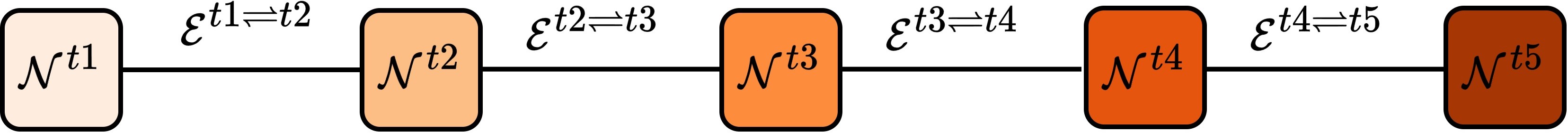}
         \caption{Adjacent}
         \label{fig:mrf_adjacent}
     \end{subfigure}
     \vspace{0.5cm}
     \vfill
     \begin{subfigure}[b]{0.45\textwidth}
         \centering
         \includegraphics[width=\textwidth]{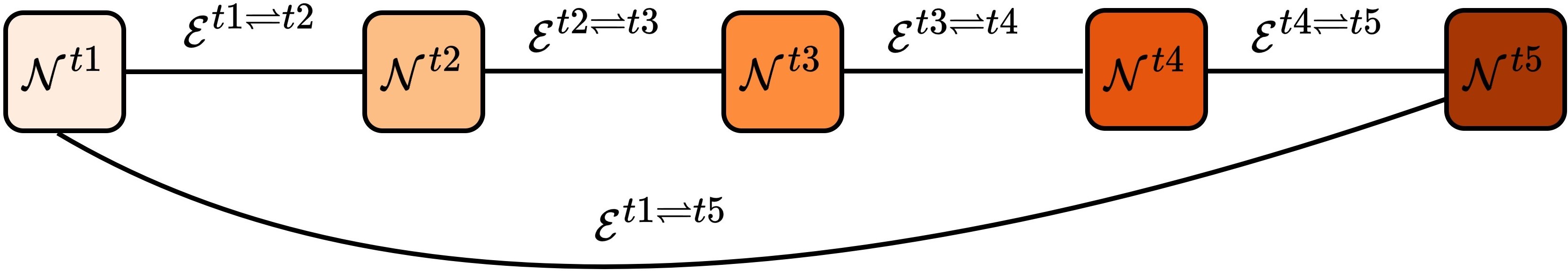}
         \caption{Cyclic}
         \label{fig:mrf_cyclic}
     \end{subfigure}
     \vspace{0.5cm}
    \vfill
     \begin{subfigure}[b]{0.45\textwidth}
         \centering
         \includegraphics[width=\textwidth]{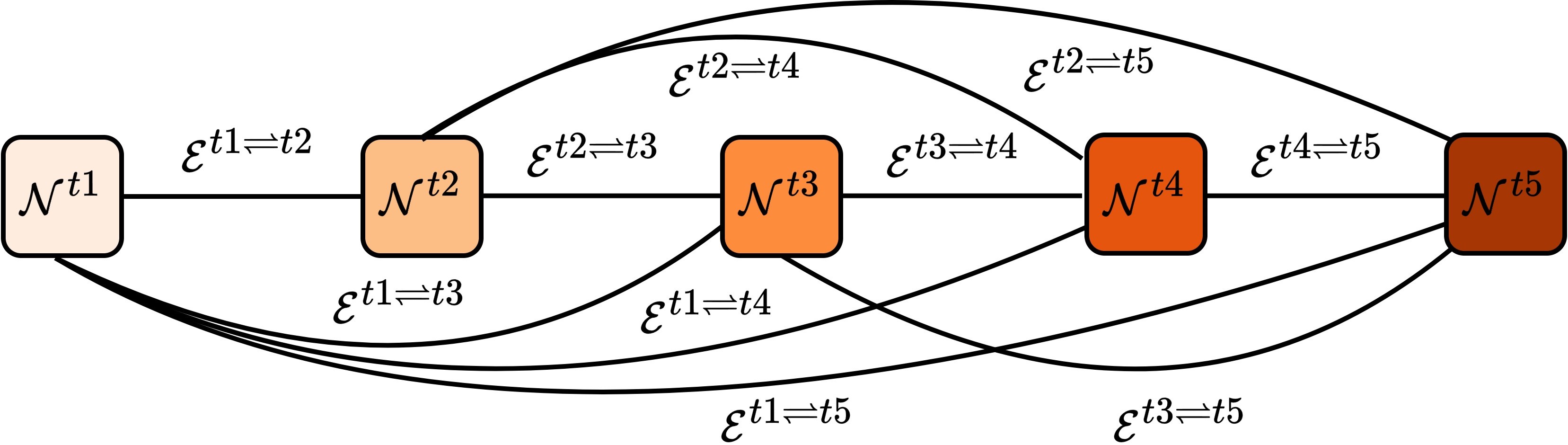}
         \caption{Dense}
         \label{fig:mrf_dense}
     \end{subfigure}
     \hfill
    \caption{Markov networks with different edge settings (exemplified for a time series with length $T = 5$). Nodes and Edges are denoted by $\mathcal{N}$ and $\mathcal{E}$, respectively.}
    \label{fig:mrf}
\end{figure}

The core idea of the MTI module is to represent segmentation and change outputs in a pairwise Markov network. This subclass of Markov networks is associated with an undirected graph $G = (\mathcal{N}, \mathcal{E})$ in which the nodes $\mathcal{N}$ correspond to random variables and the edges $\mathcal{E}$ represent pairwise relationships between the nodes (see \cite{koller2009probabilistic}). For a given pixel, we construct a Markov network with $T$ nodes corresponding to the timestamps in an image time series. Specifically, all nodes in the network correspond to a binary variable representing the presence of buildings (i.e., $\mathcal{N}^t \in \{true, false\}$). We use state 1 to denote $true$, representing the presence of a building, and state 0 to denote $false$, representing the absence of a building. Adjacent nodes in the network are connected with $N-1$ edges, where we use $\mathcal{E}^{t\rightleftharpoons k}$ to denote an edge connecting timestamps $t$ and $k$. We refer to this Markov network structure as an adjacent network (see Fig. \ref{fig:mrf_adjacent}).

To represent the segmentation and change outputs, the graph structure needs to be associated with a set of parameters that capture the relationships between nodes. The parameterization in a pairwise Markov network is achieved by assigning \textit{factors} over nodes or edges, where a factor $\phi$, also referred to as \textit{potential}, is a function from value assignments of random variables to real positive numbers $\mathbb{R}^+$. Thus, a pairwise Markov network is associated with a set of node potentials $\{\phi(\mathcal{N}_t): t = 1, ..., T \}$ and a set of edge potentials $\{\phi(\mathcal{N}_t, \mathcal{N}_k) : (\mathcal{N}_t, \mathcal{N}_k) \in G \}$. The overall distribution represented by the network is then the normalized product of all the node and edge potentials.

The segmentation outputs for a specific pixel are incorporated into the Markov network by assigning a factor $\phi_t$ over each node $\mathcal{N}_t$. Then, the segmentation outputs are encoded as node potentials, as follows:

\begin{equation}
\label{eq:node_potentials}
\begin{split}
    \phi_t(\mathcal{N}^{t} = 1) & =\mathbf{O}_{\rm seg}^{(i, j), t} \\
    \phi_t(\mathcal{N}^{t} = 0) & = 1 - \mathbf{O}_{\rm seg}^{(i, j), t}.
\end{split}
\end{equation}

These node potentials characterize the bias of nodes towards state 1 or 0, representing the presence or absence of a building, respectively. We refer to this Markov network as \textit{degenenrate} network, characterized by the absence of functions that capture the interactions between nodes.

To incorporate the change outputs for a specific pixel, we first include additional edges for the edge settings cyclic and dense. Specifically, for the cyclic case (Fig. \ref{fig:mrf_cyclic}), edge $\mathcal{E}^{t1\rightleftharpoons T}$, connecting the first node $\mathcal{N}^{t1}$ and the last node $\mathcal{N}^{T}$, is added. On the other hand, for the dense case (Fig. \ref{fig:mrf_dense}), all possible non-adjacent edges are added to the graph. Then, we define factors over the edges in the Markov network to add pairwise interactions of connected nodes. Specifically, we define pairwise potentials $\phi_{n}$ for each edge $\mathcal{E}^{n}$, connecting two nodes  $\mathcal{N}^{t}$ and $\mathcal{N}^{k}$. Since all variables in the network are binary, each factor over an edge has four parameters. The change outputs are encoded as edge potentials for the four combinations of states, as follows:

\begin{equation}
\label{eq:edge_potentials}
\begin{split}
    \phi_n(\mathcal{N}^{t} = 0,\mathcal{N}^{k} = 1) & = \mathbf{O}_{\rm ch}^{(i, j), n} \\
    \phi_n(\mathcal{N}^{t} = 1,\mathcal{N}^{k} = 0) & = \mathbf{O}_{\rm ch}^{(i, j), n} \\
    \phi_n(\mathcal{N}^{t} = 0,\mathcal{N}^{k} = 0) & = 1 - \mathbf{O}_{\rm ch}^{(i, j), n} \\
    \phi_n(\mathcal{N}^{t} = 1,\mathcal{N}^{k} = 1) & = 1 - \mathbf{O}_{\rm ch}^{(i, j), n},
\end{split}
\end{equation}

\indent where edge $n$ is connecting timestamps $t$ and $k$.

The value associated with each particular assignment of states denotes the affinity between the two states. Consequently, the higher the value assigned to the edge potential for a particular combination of states, the more compatible these two states are.

To define a global model from the local interactions defined in the parameterization of the Markov network, we take the product of the local factors and then normalize it. Once the distribution is defined, we perform a \textit{maximum a posteriori} query to find the optimal state assignment for each node in the graph. The optimal state assignment corresponds to the configuration that minimizes the overall energy determined by the node and edge potentials assigned to the graph, as defined in Equations \ref{eq:node_potentials} and \ref{eq:edge_potentials}. Therefore, the resulting state assignment for the nodes is optimal with respect to the potentials obtained from the network outputs but not necessarily with respect to the segmentation and change labels. We perform inference using the belief propagation algorithm (see Algorithm 10.4 in \cite{koller2009probabilistic}). Due to the absence of loops in the graph, belief propagation provides an exact solution. Finally, it should be noted that the MTI module does not contain any trainable parameters and is only deployed during inference.

\section{Experimental setting}\label{sec:experimental_setting}

\subsection{Datasets}
\label{subsec:datasets}

A diverse set of multi-temporal datasets is used to evaluate the proposed method. Table \ref{tab:datasets} compares key characteristics of these datasets, and detailed descriptions are provided in the following paragraphs.

\begin{table*}[!h]
\small
  \caption{Overview of dataset characteristics for SpaceNet 7, the Wuhan Urban Semantic Understanding (WUSU) dataset, and the Time Series Change Detection (TSCD) dataset.}
  \label{tab:datasets}
  \centering
  \begin{threeparttable}
  \begin{tabular}{lllccc}
    \toprule
    Dataset & Location & Satellite & \multicolumn{3}{c}{Resolution} \\
    \cmidrule{4-6}
     & & & Spectral & Spatial & Temporal \\
    \midrule
    SpaceNet 7 & Global (60 sites) & PlanetScope & 3 bands (RGB) & 4 m & 5 images\tnote{*} (2017 to 2020)\\
    WUSU & Wuhan, China & Gaofen-2 & 4 bands (RGB + NIR) & 1 m & 3 images (2015, 2016, 2018)\\
    TSCD & Chengdu, China & WorldView-2 & 3 bands (RGB) & 0.5 m & 4 images (2016, 2018, 2020, 2022) \\
    \bottomrule
  \end{tabular}
  \begin{tablenotes}
    \item[*] Approximately 24 monthly timestamps acquired between 2017 and 2020 are available for each site in the dataset.
  \end{tablenotes}
  \end{threeparttable}
\end{table*}

\textbf{SpaceNet 7}: The SpaceNet 7 dataset features time series of satellite images acquired by the PlanetScope constellation between 2017 and 2020 for 60 sites spread across the globe  \cite{van2021multi}. Each time series consists of about 24 monthly mosaics with a spatial resolution of 4 m (approximately 1024 x 1024 pixels). Furthermore, the SpaceNet 7 dataset provides manually annotated building footprints, whereas annotations are missing for image parts affected by clouds. While the task of the original SpaceNet 7 challenge was to track these building footprints (i.e., vector data), the SpaceNet 7 dataset has been leveraged for a diverse set of tasks such as urban mapping \cite{hafner2022unsupervised}, building counting \cite{razzak2023multi}, and urban change forecasting \cite{metzger2023urban}. To split the SpaceNet 7 dataset into training, validation, and test areas, we apply the within-scene splits recommended in \cite{razzak2023multi}. Specifically, approximately 80 \% of a scene (top part) is used as a source of training patches, and the remaining 20 \% (bottom part) is divided evenly into validation (bottom-left part) and test (bottom-right part) areas. Within-scene splits minimize the occurrence of out-of-distribution data during testing while simultaneously avoiding data leakage between the training and test set by utilizing spatially disjoint areas for the different dataset splits. During training, samples from the training areas are generated by randomly selecting $T$ timestamps from the time series of a site. The rasterized building labels (see \cite{hafner2022urban}) for these timestamps, were used to compute the change label. We draw 100 samples from each site during an epoch to reach an adequate number of steps before model evaluation. For model evaluation (validation and testing), the first and the last cloud-free images of a time series, in addition to evenly spaced intermediate images, were selected.

\textbf{WUSU}: The Wuhan Urban Semantic Understanding (WUSU) dataset features tri-temporal high-resolution Gaofen-2 images covering two districts in Wuhan (Hubei Province, central China) in 2015, 2016, and 2018 \cite{shi2023openWUSU}. The preprocessing workflow of the satellite images includes orthographic correction and multi-temporal image registration. Furthermore, the four multi-spectral bands acquired at 4 m spatial resolution are pansharpened to a spatial resolution of 1 m, resulting in images of size 6358 x 6382 and 7025 x 5500 pixels for Hongshan District and Jiang'an District, respectively. In addition to the Gaofen-2 images, the WUSU dataset provides corresponding land-use/land-cover (LULC) labels, including manually refined building annotations (Class 2 Low building and Class 3 High building). Since the proposed method requires binary building labels, Class 2 and Class 3 were remapped to the foreground class, whereas all other classes were remapped to the background class. We follow the within-scene split recommended by the authors, using the top halves of the six images for the test set and the bottom halves for the training set that was further divided into training (90 \%) and validation (10 \%) tiles.

\textbf{TSCD}: The Time Series Change Detection (TSCD) dataset features bi-annual WorldView-2 satellite images acquired over Chengdu (Sichuan Province) between 2016 and 2022 \cite{zhao2024coud}. The images have a resolution of approximately 0.5 m and are split into 512 x 512 pixel tiles. The tiles are divided into a training, validation, and test set. The TSCD dataset provides building change labels for each adjacent image pair (2016--2018, 2018--2020, 2020--2022). We derived change labels for an arbitrary image pair from the time series by considering all adjacent change labels connecting this pair and computing the number of changes. An odd number of adjacent changes indicates change between the image pair, whereas an even number indicates no change.

\subsection{Baseline and benchmark methods}

We selected a comprehensive set of baseline and benchmark methods for quantitative and qualitative comparisons with the proposed methods. These selected methods are grouped into two categories, which are listed below.

\textbf{Bi-temporal change detection methods}

\setlist[enumerate]{start=1}
\begin{enumerate}[label=\arabic*.,ref=\arabic*]
    \item Siam-Diff \cite{daudt2018fully} employs a Siamese encoder to extract feature maps from bi-temporal images. A decoder produces the change prediction from the subtracted feature maps. The encoders and decoder follow the U-Net architecture. 
    \item SNUNet \cite{fang2021snunet} replaces the architecture in the Siam-Diff network with a Nested U-Net (UNet++ \cite{zhou2018unet++}). In addition, a channel attention module is incorporated into the architecture.
    \item DTCDSCN \cite{liu2020building} combines a typical Siamese ConvNet for bi-temporal change detection with a dual attention module and two additional decoders with shared weights for building segmentation.
    \item BIT \cite{chen2021remote} employs a bi-temporal image transformer module that operates in a compact token space to refine features extracted by a Siamese ConvNet.  
    \item AMTNet \cite{liu2023attention} also extracts features using a Siamese ConvNet, and combines attention mechanisms and multi-scale processing techniques to model contextual information in bi-temporal images.
    \item ScratchFormer \cite{noman2024remote} introduces shuffled sparse attention layers in a Siamese ConvNet encoder to effectively capture semantic changes when training from scratch.

\end{enumerate}

\textbf{Multi-temporal change detection/segmentation methods}
\setlist[enumerate]{start=7}
\begin{enumerate}[label=\arabic*.,ref=\arabic*]
    \item L-UNet \cite{papadomanolaki2019detecting} employs a shared U-Net for multi-scale feature extraction in SITS and uses LSTM modules \cite{hochreiter1997long} for temporal modeling. The LSTM modules produce a single multi-scale feature map which is transformed into the output feature map using a U-Net decoder.
    \item Multi-Task L-UNet \cite{papadomanolaki2021deep} adds a semantic U-Net decoder to L-UNet to segment buildings in the first and last images of a time series.
    \item U-TAE \cite{garnot2021panoptic} uses a shared U-Net encoder to extract multi-scale feature maps for the SITS. A temporal attention encoder (L-TAE \cite{garnot2020lightweight} is then used to collapse the temporal dimension, before using the U-Net decoder to produce a single output feature map.
    \item TSViT \cite{tarasiou2023vits}  splits a SITS into non-overlapping patches in space and time which are tokenized and subsequently processed by a temporo-spatial encoder. A segmentation head reassembles the encoded features into a single output feature map.  
    \item U-TempoNet \cite{cai2024cost} uses a shared ConvNet encoder to extract multi-scale feature maps for all images. Subsequently, a single multi-scale feature map, obtained through temporal modeling with a bidirectional LSTM, is processed with a decoder to produce the output feature map.
\end{enumerate}

The Siamese ConvNets Siam-Diff, SNUNet, and DTCDSCN are commonly used as change detection baselines, whereas BIT, AMTNet, and ScratchFormer represent recent methods combining Siamese ConvNets with transformers. On the other hand, L-UNet and Multi-Task L-UNet are benchmark methods for multi-temporal change detection.It should be noted that Multi-Task L-UNet and DTCDSCN are multi-task methods that perform change detection and building segmentation. Finally, U-TAE, TSViT and U-TempoNet are recent segmentation methods for SITS inputs that can be adopted for multi-temporal change detection without architectural changes.

\subsection{Model evaluation}

Three accuracy metrics were used for the quantitative assessment of model predictions: F1 score, intersection over union (IoU), and overall accuracy (OA). Formulas for the metrics are given below (Equations \ref{eq:f1_score}, \ref{eq:iou}, and \ref{eq:oa}), where TP, TN, FP, and FN represent the number of true positive, true negative, false positive, and false negative pixels, respectively.

\begin{equation}
\label{eq:f1_score}
    \text{F1 score} = \frac{TP}{TP + \frac{1}{2}(FP + FN)}
\end{equation}

\begin{equation}
\label{eq:iou}
    \text{IoU} = \frac{TP}{TP + FP + FN},
\end{equation}

\begin{equation}
\label{eq:oa}
    \text{OA} = \frac{TP + TN}{TP + TN + FP + FN},
\end{equation}

Using these two accuracy metrics, we assessed model performance across three tasks to accommodate the large variety of baseline and benchmark methods. These tasks are described in detail in the following:

\begin{itemize}
    \item \textit{Bi-temporal change detection} measures the accuracy of the predicted changes between the first and last image of a time series.
    \item \textit{Continuous change detection} measures the average accuracy of the predicted changes between consecutive image pairs in a time series.
    \item \textit{Segmentation} measures the accuracy of the building predictions corresponding to the last image of a time series.
\end{itemize} 

The first task focuses on urban change detection from image pairs acquired multiple years apart. This task is considered by most urban change detection methods. For bi-temporal change detection methods, changes were directly predicted based on the first-last image pair, ignoring intermediate images in a time series. On the other hand, the second task focuses on assessing change predictions between consecutive image pairs in a time series. Consequently, the continuous urban change detection task focuses on image pairs with periods between acquisition dates that are considerably shorter (i.e., annual and sub-annual). The last task assesses the auxiliary segmentation task of multi-task methods and segmentation models. It should be noted that the change detection performance of segmentation models is not assessed because post-classification comparison suffers from the accumulation of classification errors \cite{liu2019learning}.

\subsection{Implementation details}
\label{subsec:implementation_details}

We implement the proposed method using the deep learning framework PyTorch \cite{paszke2019pytorch}. In addition, the einops package \cite{rogozhnikov2022einops} was used to efficiently reshape feature maps, and the pgmpy package \cite{Ankan2015} to implement the Markov network and perform belief propagation.
Models were trained for a maximum duration of 100 epochs on NVIDIA GeForce RTX 3080 graphics cards, using early stopping with patience 10 to prevent models from overfitting to the training set. AdamW was used as optimizer \cite{loshchilov2018fixing} with a linear learning rate scheduler. The remainder of this section describes the training setup in detail.

\textbf{Augmentations}: To enhance the training dataset, we applied four data augmentation operations, namely rotations ($k \cdot 90^{\circ}$, where $k$ is randomly selected from $\{0, 1, 2, 3\}$), flips (horizontal and vertical with a probability of 50 \%), Gaussian blur, and random color jittering. The parameters that determine how much to jitter the brightness, contrast, saturation, and hue of an image were set to 0.3 \cite{bandara2022transformer}. For validation and testing, on the other hand, no data augmentation was applied.

\textbf{Oversampling}: To account for the fact that the occurrence of change is usually considerably less frequent than no change \cite{bovolo2015time}, change areas were oversampled during network training. For a given site, twenty patches of size \(64 \times 64\) pixels were randomly cropped from the change label, before assigning each patch a probability according to its change pixel percentage, including a base probability for patches with no change pixels. A single patch was chosen based on those probabilities. For transformer-based methods (BIT, AMTNet, and Scratchformer), the patch size was increased to \(128 \times 128\) pixels to include more long-range spatial context.

\textbf{Hyper-parameter tuning}: For each model, hyper-parameters were tuned empirically on the validation set using grid search. Specifically, an exhaustive search with three learning rates ($1 \cdot 10^{-5}$, $5 \cdot 10^{-5}$, $1 \cdot 10^{-4}$) and two batch sizes (8, 16) was performed to determine the optimum values of hyper-parameters. Then, five models were trained with the best hyper-parameters but different seeds for weight initialization and data shuffling. Consequently, reported values correspond to the average of five runs.

Multi-Task L-UNet requires additional hyper-parameters to balance the contribution of the segmentation and change loss terms, which we adopted from the paper \cite{papadomanolaki2021deep}. All bi-temporal urban change detection methods were trained on the cyclic edges setting (see Fig. \ref{fig:mrf_cyclic}).

\section{Experiment results}

In this section, we present the quantitative and qualitative results on the SpaceNet 7 and WUSU datasets, and the ablation study results. It should be noted that all accuracy values are reported on the respective test sets and correspond to the mean values obtained from five models. These were trained with different seeds using the best hyper-parameters determined with a grid search (see Section \ref{subsec:implementation_details}). On the other hand, the median model is used for the qualitative results, which are only shown for competitive methods selected based on the quantitative results.

\subsection{SpaceNet 7}

The quantitative results for the SpaceNet 7 dataset are listed in Table \ref{tab:quan_results_sn7}. The proposed method achieved the highest F1 scores and IoU values for both urban change detection tasks (i.e., bi-temporal and continuous). Several multi-temporal models (L-UNet, Multi-Task L-UNet, and U-TempoNet) outperform bi-temporal change detection methods on the bi-temporal task while others are less effective (U-TAE and TSViT). Among the bi-temporal methods, ScartchFormer and the ConvNet-based methods Siam-Diff and SNUNet achieved the highest accuracy values. For building segmentation, the proposed method also outperformed the other multi-task methods including DTCDSCN and Multi-Task L-UNet.

\begin{table*}[!h]
\small
  \caption{Quantitative results on the SpaceNet 7 test areas. The best and second-best performances are highlighted in red and blue, respectively. "-" denotes that the accuracy metric does not apply to a specific method since the corresponding variable is not predicted.}
  \label{tab:quan_results_sn7}
  \centering
  \begin{tabular}{lccccccc}
    \toprule
    \multirow{3}{*}{Method} & \multicolumn{2}{c}{Change detection} & Segmentation \\
     & Bi-temporal & Continuous & \\
    & (F1 / IoU / OA) & (F1 / IoU / OA) & (F1 / IoU / OA) \\
    \midrule
    Siam-Diff & 0.453 / 0.293 / 98.8 & 0.273 / 0.158 / 99.5 & - \\
    SNUNet & 0.454 / 0.294 / 98.8 & 0.300 / 0.177 / 99.6 & - \\
    DTCDSCN & 0.413 / 0.260 / 98.7 & 0.250 / 0.143 / 99.6 & 0.488 / 0.323 / 92.3 \\
    BIT & 0.386 / 0.239 / 99.0 & 0.275 / 0.160 / 99.6 & - \\
    AMTNet & 0.424 / 0.269 / 98.7 & 0.282 / 0.164 / 99.6 & - \\
    ScratchFormer& 0.468 / 0.305 / 98.9 & \textcolor{blue}{0.328 / 0.196 / 99.6} & - \\
     \cmidrule{1-7}
    L-UNet & \textcolor{blue}{0.519 / 0.350 / 98.9} & - & - \\
    MT L-UNet & 0.515 / 0.347 / 98.9 & - & \textcolor{blue}{0.512 / 0.344 / 92.7} \\
    U-TAE & 0.366 / 0.225 / 97.5 & - & - \\
    TSViT & 0.168 / 0.092 / 97.2 & - & - \\
    U-TempoNet & 0.494 / 0.328 / 98.8 & - & - \\
     \cmidrule{1-7}
    Proposed  & \textcolor{red}{0.551 / 0.381 / 99.0} & \textcolor{red}{0.414 / 0.261 / 99.7} & \textcolor{red}{0.596 / 0.424 / 94.3}  \\
    \bottomrule
  \end{tabular}
\end{table*}

\begin{figure*}[!h]
\centering
\includegraphics[width=\textwidth]{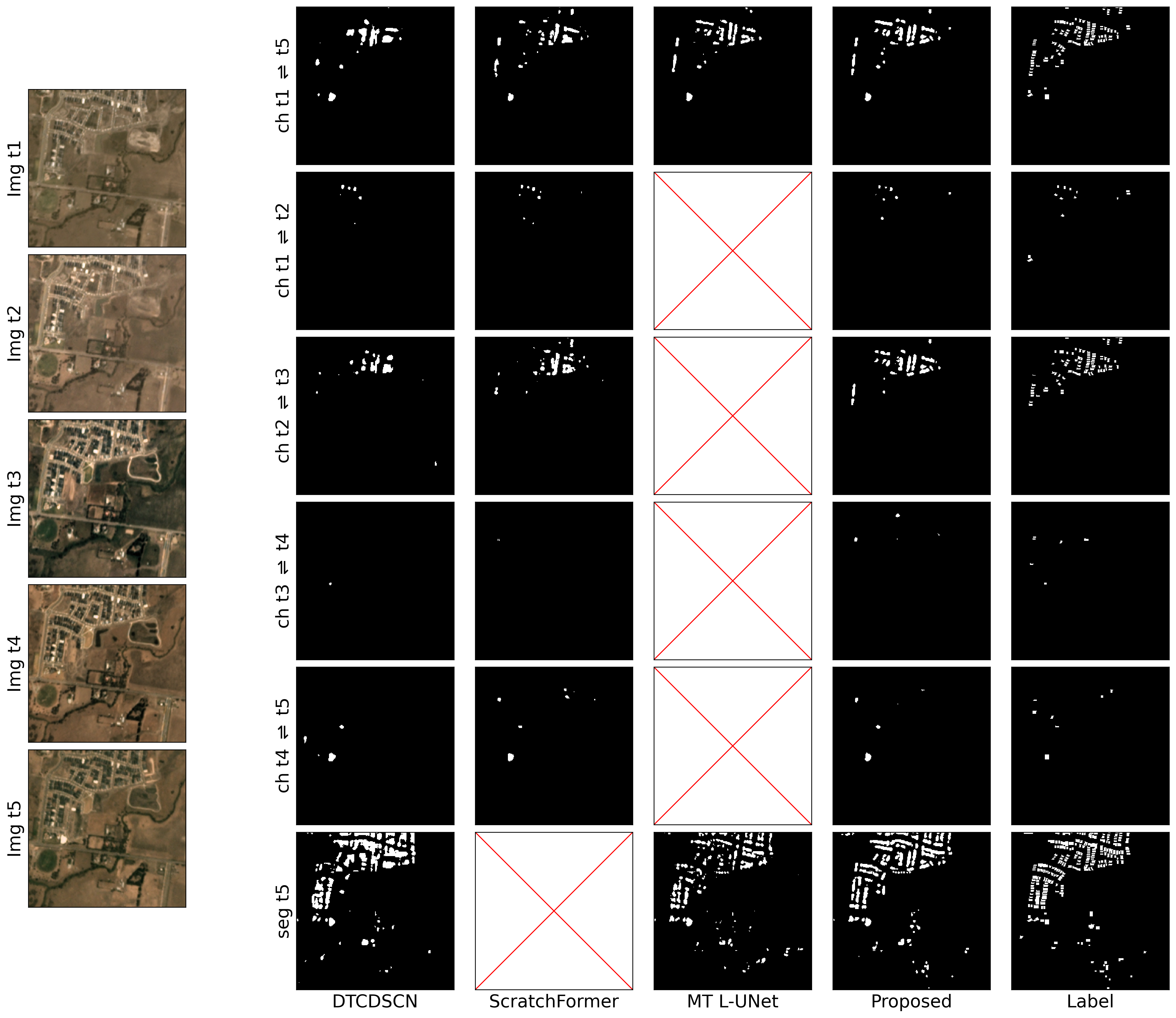}
\caption{Qualitative urban change and building segmentation results for a SpaceNet 7 test site located in Australia. The PlanetScope satellite image time series is shown on the left, and the model outputs and the label are shown on the right. The top row shows the changes between the first and last image of the time series, rows two to five show the continuous changes between consecutive image pairs, and the bottom row the buildings segmentation corresponding to the last image.}
\label{fig:qual_results_sn7_1}
\end{figure*}

\begin{figure*}[!h]
\centering
\includegraphics[width=\textwidth]{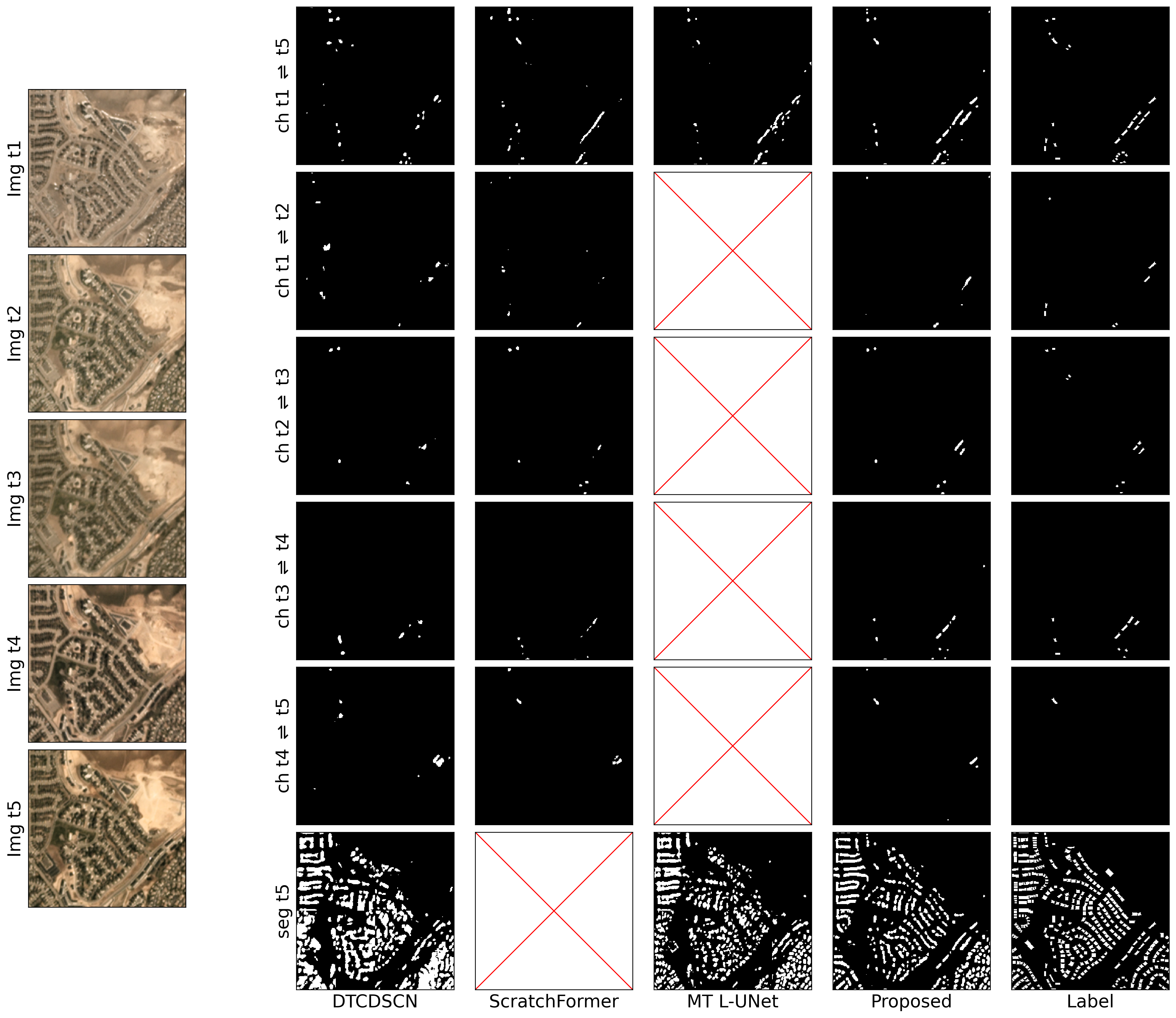}
\caption{Qualitative urban change and building segmentation results for a SpaceNet 7 test site located in the United States. The PlanetScope satellite image time series is shown on the left, and the model outputs and the label are shown on the right. The top row shows the changes between the first and last image of the time series, rows two to five show the continuous changes between consecutive image pairs, and the bottom row the buildings segmentation corresponding to the last image.}
\label{fig:qual_results_sn7_2}
\end{figure*}

Fig. \ref{fig:qual_results_sn7_1} and \ref{fig:qual_results_sn7_2} show qualitative change detection and building segmentation results for two SpaceNet 7 test sites located in Australia and the United States, respectively. For both sites, the proposed method detects urban changes more accurately than competing methods (DTCDSCN, ScratchFormer, and Multi-Task L-UNet). In particular, the continuous change detection results for consecutive image pairs (rows two to five) show a better agreement with the label than those of the other methods. In addition, the change outputs of the proposed method show a high level of consistency, meaning that the aggregated continuous changes correspond to the bi-temporal changes between the first and the last image (top row). Finally, the proposed method maps buildings with more detail than the competing methods, as shown in the bottom row of Fig. \ref{fig:qual_results_sn7_1} and \ref{fig:qual_results_sn7_2}.

\subsection{WUSU}

Change detection performance on the WUSU dataset is lower than on the SpaceNet 7 dataset (Table \ref{tab:quan_results_wusu}). For bi-temporal change detection, only ScratchFormer, L-UNet, and Multi-Task L-UNet exceed an F1 score of 0.275 and an IoU value of 0.160. In comparison, accuracy values for the continuous change detection task are slightly higher, except for the proposed method. Overall, our method outperformed all other methods on both change detection tasks. This also applied to the building segmentation task, where the proposed method achieved an F1 score of 0.663 and an IoU value of 0.496.

\begin{table*}[!h]
\small
  \caption{Quantitative results on the WUSU test areas. The best and second-best performances are highlighted in red and blue, respectively. "-" denotes that the accuracy metric does not apply to a specific method since the corresponding variable is not predicted.}
  \label{tab:quan_results_wusu}
  \centering
  \begin{tabular}{lccccccc}
    \toprule
    \multirow{3}{*}{Method} & \multicolumn{2}{c}{Change detection} & Segmentation \\
     & Bi-temporal & Continuous & \\
    & (F1 / IoU / OA) & (F1 / IoU / OA) & (F1 / IoU / OA) \\
    \midrule
    Siam-Diff & 0.175 / 0.096 / 96.1 & 0.236 / 0.134 / 97.4 & - \\
    SNUNet & 0.188 / 0.104 / 94.9 & 0.211 / 0.118 / 96.5 & - \\
    DTCDSCN & 0.278 / 0.162 / 96.2 & 0.318 / 0.189 / 97.6 & \textcolor{blue}{0.539 / 0.369 / 84.2} \\
    BIT & 0.213 / 0.120 / 96.5 & 0.314 / 0.187 / 97.9 & - \\
    AMTNet & 0.187 / 0.104 / 95.9 & 0.264 / 0.152 / 97.6 & - \\
    ScratchFormer & \textcolor{blue}{0.324 / 0.193 / 96.6} & \textcolor{blue}{0.352 / 0.214 / 97.8} & - \\
     \cmidrule{1-7}
    L-UNet & 0.279 / 0.162 / 96.0 & - & - \\
    MT L-UNet & 0.276 / 0.161 / 96.1 & - & 0.479 / 0.315 / 83.1 \\
    U-TAE & 0.267 / 0.154 / 92.1 & - & - \\
    TSViT & 0.219 / 0.123 / 92.1 & - & - \\
    U-TempoNet & 0.246 / 0.141 / 95.0 & - & - \\
     \cmidrule{1-7}
    Proposed  & \textcolor{red}{0.440 / 0.282 / 97.0} & \textcolor{red}{0.389 / 0.242 / 98.3} & \textcolor{red}{0.663 / 0.496 / 88.9}  \\
    \bottomrule
  \end{tabular}
\end{table*}

\begin{figure*}[!h]
\centering
\includegraphics[width=\textwidth]{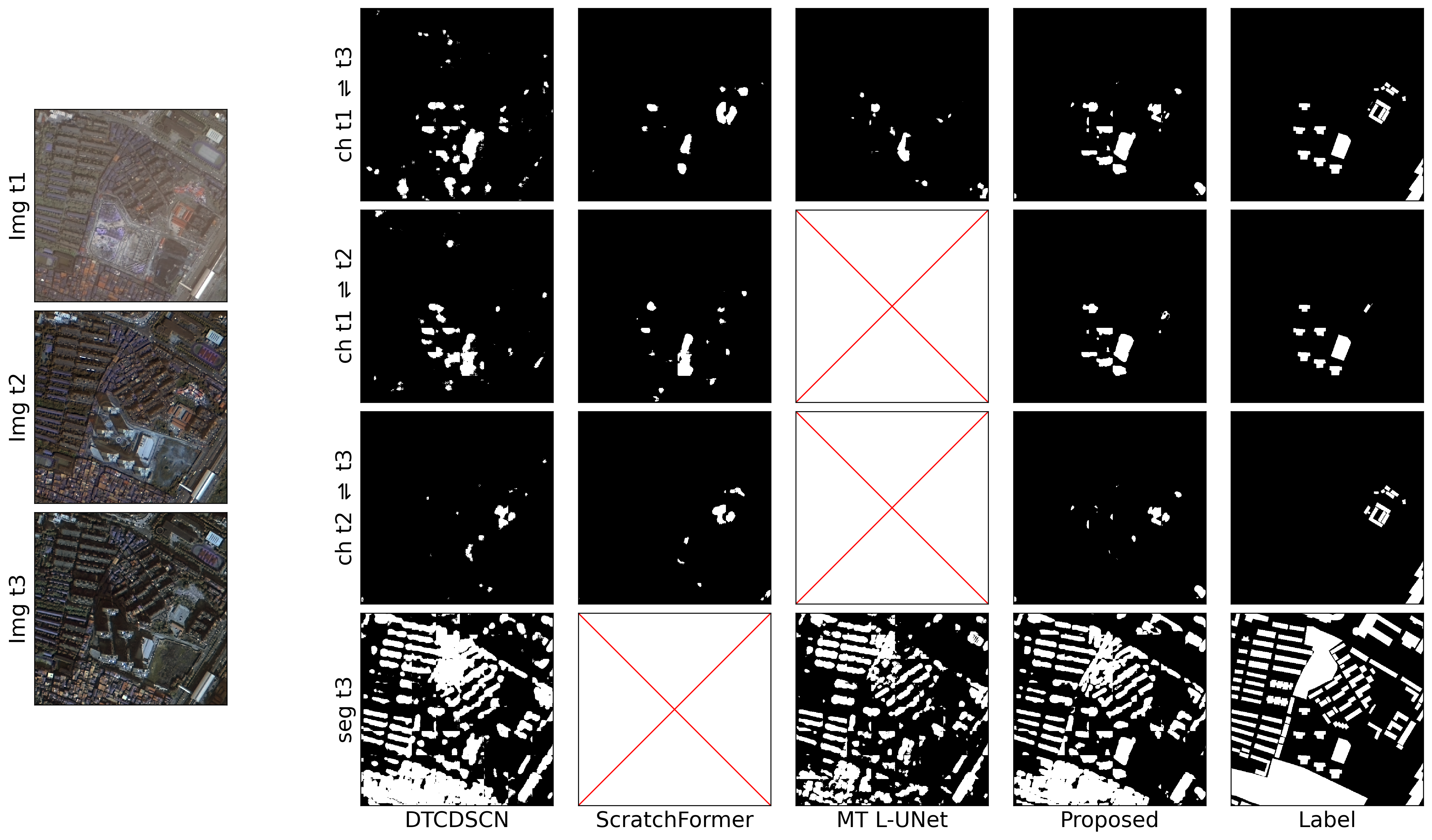}
\caption{Qualitative urban change and building segmentation results for a WUSU test site located in Wuhan's Jiang’an
District, China. The Gaofen-2 satellite image time series is shown on the left, and the model outputs and the label are shown on the right. The top row shows the changes between the first and last image of the time series, rows two and three show the continuous changes between consecutive image pairs, and the bottom row the buildings segmentation corresponding to the last image.}
\label{fig:qual_results_wusu_1}
\end{figure*}

\begin{figure*}[!h]
\centering
\includegraphics[width=\textwidth]{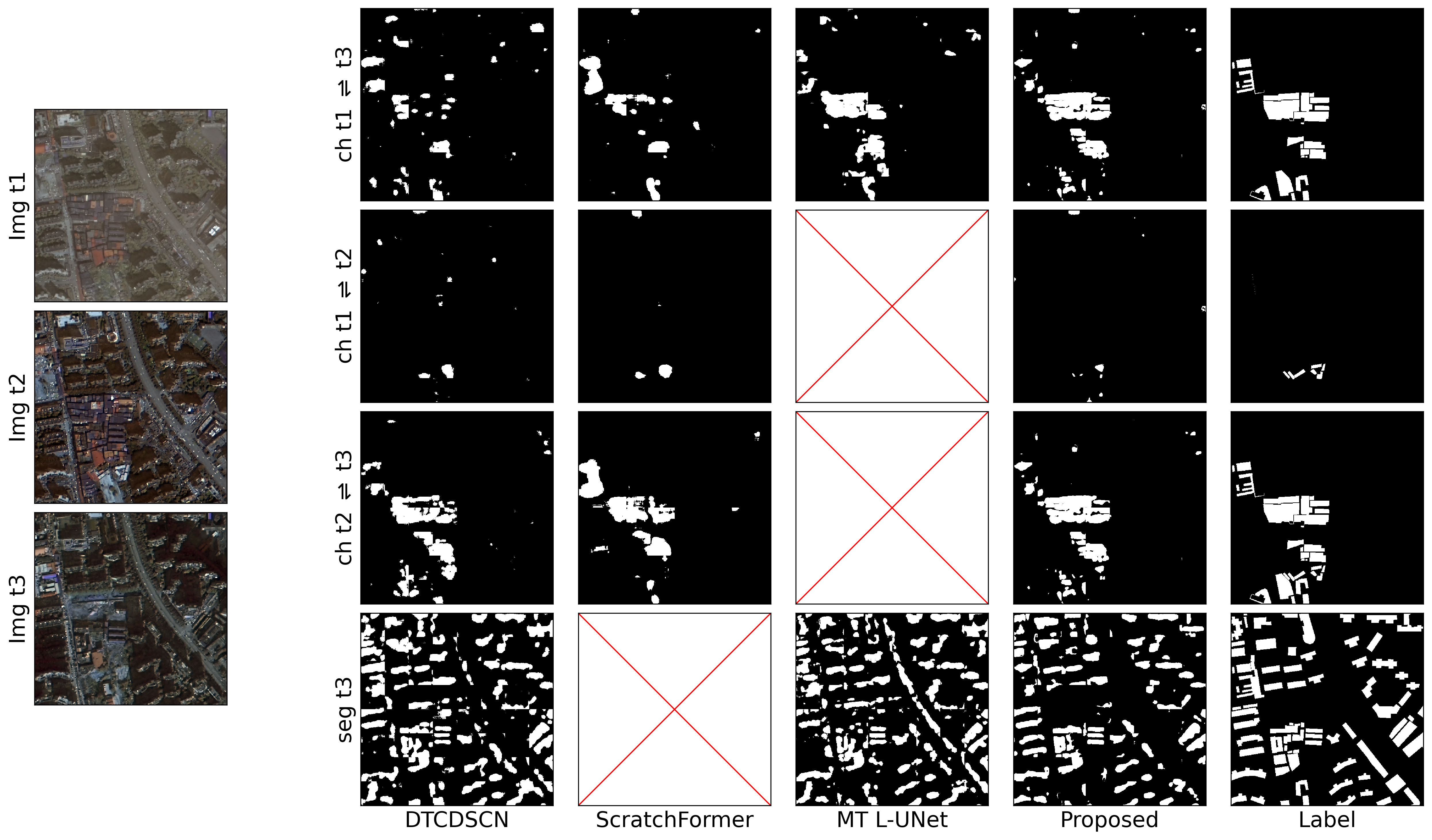}
\caption{Qualitative urban change and building segmentation results for a WUSU test site located in Wuhan's Jiang’an
District, China. The Gaofen-2 satellite image time series is shown on the left, and the model outputs and the label are shown on the right. The top row shows the changes between the first and last image of the time series, rows two and three show the continuous changes between consecutive image pairs, and the bottom row the buildings segmentation corresponding to the last image.}
\label{fig:qual_results_wusu_2}
\end{figure*}

Fig. \ref{fig:qual_results_wusu_1} and \ref{fig:qual_results_wusu_2} show qualitative change detection and building segmentation results for two WUSU test sites located in Wuhan's Jiang'an District in China. The change detection results produced by the proposed method show good agreement with the label, especially in comparison with the competing methods. As for SpaceNet 7, the proposed method achieved a high level of consistency between the continuous change detection outputs (rows two and three) and the change output between the first and last image (top row). In contrast, the bi-temporal change detection methods identified changes between the first and last images (ch $t1 \rightleftharpoons t3$) that are present in neither of the continuous change rows (ch $t1 \rightleftharpoons t2$ and ch $t2 \rightleftharpoons t3$). Finally, the bottom row demonstrates that all methods accurately map buildings, but the proposed method achieved more detailed building delineations than its competitors. It should also be noted that the label does not distinguish individual buildings in very dense built-up areas (e.g., the bottom left area in Fig. \ref{fig:qual_results_wusu_1}).

\subsection{TSCD}

The quantitative results for the TSCD dataset are listed in Table \ref{tab:quan_results_tscd}. Multi-temporal methods outperformed bi-temporal methods on detecting changes between the first and last timestamps of SITS. On the continuous task, bi-temporal methods capable of modeling long-range spatial dependencies in VHR imagery (BIT, AMTNet, and ScratchFormer) outperformed Siam-Diff and SNUNet. However, our method achieved the highest performance on both tasks despite not leveraging spatial attention.  

\begin{table}[!h]
\small
  \caption{Quantitative results on the TSCD test areas. The best and second-best performances are highlighted in red and blue, respectively. "-" denotes that the accuracy metric does not apply to a specific method since the corresponding variable is not predicted.}
  \label{tab:quan_results_tscd}
  \centering
  \begin{tabular}{lcc}
    \toprule
    \multirow{3}{*}{Method} & \multicolumn{2}{c}{Change detection} \\
     & Bi-temporal & Continuous \\
    & (F1 / IoU / OA) & (F1 / IoU / OA) \\
    \midrule
    Siam-Diff & 0.211 / 0.118 / 89.3 & 0.284 / 0.166 / 93.2 \\
    SNUNet & 0.220 / 0.123 / 89.4 & 0.328 / 0.197 / 92.6 \\
    BIT & 0.203 / 0.114 / 90.0 & 0.363 / 0.222 / 94.5 \\
    AMTNet & 0.298 / 0.178 / 91.1 & 0.398 / 0.249 / 94.9 \\
    ScratchFormer & 0.270 / 0.157 / 91.7 & \textcolor{blue}{0.504 / 0.340 / 95.9} \\
     \midrule
    L-UNet & 0.401 / 0.251 / 84.6 & - \\
    U-TAE & 0.415 / 0.262 / 81.6 & - \\
    TSViT & 0.355 / 0.216 / 79.3 & - \\
    U-TempoNet & \textcolor{blue}{0.435 / 0.279 / 85.1} & - \\
     \midrule
    Proposed  & \textcolor{red}{0.543 / 0.373 / 93.7} & \textcolor{red}{0.573 / 0.402 / 96.6} \\
    \bottomrule
  \end{tabular}
\end{table}

\subsection{Ablation study}

\textbf{Loss function}: Table \ref{tab:ablation_loss} shows the ablation results for different change loss edge settings (first-last, adjacent, cyclic, and dense) and the segmentation loss. Multi-task integration was disabled for this experiment to isolate the effect of the loss function on network performance. It should also be noted that the settings adjacent, cyclic, and dense all require building annotations for each image in the time series as labels (see Section \ref{subsec:loss_function}), whereas first-last only requires building annotations for the first and last images. Considering additional change edges in the loss function generally improves continuous change detection performance. In contrast, a single change loss term suffices for the bi-temporal change detection task. The segmentation loss term improves performance for both change detection tasks, which holds for all change loss settings. Therefore, the optimal loss setting consists of a change loss with dense edges combined with a segmentation loss.

\begin{table*}[!h]
\small
  \caption{Ablation results for the loss function with different edge settings for the change loss term. The best and second-best performances on the SpaceNet 7 (SN7), WUSU, and TSCD datasets are highlighted in red and blue, respectively.}
  \label{tab:ablation_loss}
  \centering
  \begin{tabular}{llcccc}
    \toprule
    \multirow{3}{*}{Dataset} & \multirow{3}{*}{Change loss} & \multirow{3}{*}{Seg loss} & \multicolumn{2}{c}{Change detection} & Segmentation \\
    & & & Bi-temporal & Continuous & \\
    & & & (F1 / IoU / OA) & (F1 / IoU / OA) & (F1 / IoU / OA) \\
    \midrule
    \multirow{8}{*}{\rotatebox[origin=c]{90}{SN7}} & first-last & \xmark & 0.453 / 0.294 / 98.7 & 0.253 / 0.145 / 99.6 & - \\
    & first-last & \cmark & 0.519 / 0.350 / 98.8 & 0.323 / 0.193 / 99.7 & 0.565 / 0.394 / 93.8 \\
    & adjacent & \xmark & 0.479 / 0.315 / 98.8 & 0.357 / 0.217 / 99.6 & - \\
    & adjacent & \cmark & 0.520 / 0.352 / 98.9 & \textcolor{blue}{0.384 / 0.238 / 99.6} & \textcolor{blue}{0.584 / 0.413 / 94.1} \\
     & cyclic & \xmark & 0.516 / 0.348 / 98.8 & 0.361 / 0.220 / 99.6 & -  \\
     & cyclic & \cmark & \textcolor{blue}{0.532 / 0.363 / 98.9} & 0.377 / 0.233 / 99.7 & 0.581 / 0.409 / 94.0 \\
     & dense & \xmark & 0.511 / 0.343 / 98.8 & 0.364 / 0.223 / 99.6 & - \\
     & dense & \cmark & \textcolor{red}{0.537 / 0.367 / 98.8} & \textcolor{red}{0.397 / 0.248 / 99.7} & \textcolor{red}{0.593 / 0.422 / 94.3} \\
    \midrule
    \multirow{6}{*}{\rotatebox[origin=c]{90}{WUSU}}
     & first-last & \xmark & 0.274 / 0.159 / 95.7 & 0.234 / 0.133 / 96.3 & -  \\
     & first-last & \cmark & \textcolor{blue}{0.401 / 0.251 / 96.5} & 0.297 / 0.175 / 98.3 & 0.650 / 0.482 / 88.4 \\    
    & adjacent & \xmark & 0.251 / 0.144 / 96.4 & 0.297 / 0.175 / 97.9 & - \\
    & adjacent & \cmark & 0.356 / 0.217 / 95.3 & \textcolor{blue}{0.373 / 0.229 / 98.2} & \textcolor{blue}{0.650 / 0.482 / 88.6} \\
     & cyclic & \xmark & 0.328 / 0.196 / 96.0 & 0.289 / 0.169 / 98.2 & - \\
     & cyclic & \cmark & \textcolor{red}{0.420 / 0.266 / 96.6} & \textcolor{red}{0.391 / 0.243 / 98.3} & \textcolor{red}{0.660 / 0.493 / 88.8}  \\
    \midrule
    \multirow{4}{*}{\rotatebox[origin=c]{90}{TSCD}}
     & first-last & \xmark & 0.270 / 0.157 / 84.9 & 0.157 / 0.086 / 62.8 \\ 
    & adjacent & \xmark & 0.255 / 0.147 / 89.6 & 0.453 / 0.296 / 95.1 & - \\
     & cyclic & \xmark & \textcolor{blue}{0.512 / 0.344 / 92.9} & \textcolor{blue}{0.553 / 0.383 / 96.4} & - \\
     & dense & \xmark & \textcolor{red}{0.543 / 0.373 / 93.7} & \textcolor{red}{0.573 / 0.402 / 96.6} & - \\
    \bottomrule
  \end{tabular}
\end{table*}

\textbf{TFR module}: We perform an additional ablation experiment investigating the contribution of the TFR module and testing if recurrent sequence models, particularly recurrent neural networks (RNNs) \cite{elman1990finding} and LSTMs \cite{hochreiter1997long}, can be considered as alternatives to self-attention. We run all settings with and without the MTI module due to the complementary nature of the modules. Table \ref{tab:ablation_tfr} shows that adding the TFR module to our framework achieves large performance gains across all tasks and datasets. Among the sequence models, self-attention outperformed the recurrent sequence models on the segmentation task across all datasets and the continuous urban change detection task on SpaceNet 7 and the TSCD dataset. It only fell short of LSTMs on the WUSU dataset. Finally, for bi-temporal change detection, self-attention outperformed RNNs on all datasets but LSTMs achieved better results before multi-task integration. In general, self-attention is the most effective sequence model, especially in combination with the MTI module. However, LSTMs should be considered for multi-temporal change detection and datasets with few timestamps.

\begin{table*}[!h]
\small
  \caption{Ablation results for the TFR module with different sequence models. The maximum number of edges was used for each experiment in the change loss and MTI module. The best and second-best performances on the SpaceNet 7 (SN7), WUSU, and TSCD datasets are highlighted in red and blue, respectively.}
  \label{tab:ablation_tfr}
  \centering
  \begin{tabular}{lclcccc}
    \toprule
    \multirow{3}{*}{Dataset} &\multirow{3}{*}{TFR module} & \multirow{3}{*}{Sequence model} & \multirow{3}{*}{MTI module} & \multicolumn{2}{c}{Change detection} & Segmentation \\
    & & & & Bi-temporal & Continuous & \\
    & & & & (F1 / IoU / OA) & (F1 / IoU / OA) & (F1 / IoU / OA) \\
    \midrule
    \multirow{8}{*}{\rotatebox[origin=c]{90}{SN7}} & \xmark & - & \xmark & 0.490 / 0.324 / 98.8 & 0.314 / 0.186 / 99.6 & 0.549 / 0.379 / 93.6  \\
     & \xmark & - & \cmark & 0.511 / 0.343 / 98.9 & 0.344 / 0.208 / 99.6 & 0.576 / 0.404 / 94.2 \\
     & \cmark & Recurrent (RNN) & \xmark & 0.523 / 0.354 / 98.8 & 0.333 / 0.200 / 99.6 & 0.556 / 0.386 / 93.5 \\
     & \cmark & Recurrent (RNN) & \cmark & 0.525 / 0.356 / 98.9 & 0.377 / 0.232 / 99.6 & 0.572 / 0.401 / 93.8 \\
     & \cmark & Recurrent (LSTM) & \xmark & \textcolor{blue}{0.549 / 0.378 / 98.9} & 0.351 / 0.213 / 99.7 & 0.571 / 0.400 / 94.1 \\
     & \cmark & Recurrent (LSTM) & \cmark & 0.547 / 0.377 / 99.0 & \textcolor{blue}{0.402 / 0.252 / 99.6} & 0.588 / 0.417 / 94.4  \\
     & \cmark & Self-attention & \xmark & 0.537 / 0.367 / 98.8 & 0.397 / 0.248 / 99.7 & \textcolor{blue}{0.593 / 0.422 / 94.3} \\
     & \cmark & Self-attention & \cmark & \textcolor{red}{0.551 / 0.381 / 99.0} & \textcolor{red}{0.414 / 0.261 / 99.7} & \textcolor{red}{0.596 / 0.424 / 94.3}  \\
    \midrule
    \multirow{8}{*}{\rotatebox[origin=c]{90}{WUSU}} & \xmark & - & \xmark & 0.342 / 0.208 / 96.8 & 0.364 / 0.223 / 97.9 & 0.583 / 0.412 / 86.2  \\
     & \xmark & - & \cmark & 0.392 / 0.245 / 96.6 & 0.339 / 0.205 / 97.6 & 0.649 / 0.480 / 88.4 \\
     & \cmark & Recurrent (RNN) & \xmark & 0.423 / 0.268 / 96.7 & 0.391 / 0.243 / 98.2 & 0.613 / 0.443 / 87.1  \\
     & \cmark & Recurrent (RNN) & \cmark & 0.440 / 0.282 / 96.9 & 0.390 / 0.243 / 98.2 & 0.658 / 0.491 / 88.4  \\
     & \cmark & Recurrent (LSTM) & \xmark & 0.426 / 0.271 / 96.8 & \textcolor{red}{0.394 / 0.245 / 98.3} & 0.615 / 0.444 / 87.3  \\
     & \cmark & Recurrent (LSTM) & \cmark & \textcolor{red}{0.441 / 0.283 / 96.9} & 0.387 / 0.240 / 98.2 & 0.658 / 0.490 / 88.5  \\
     & \cmark & Self-attention & \xmark & 0.420 / 0.266 / 96.6 & \textcolor{blue}{0.391 / 0.243 / 98.3} & \textcolor{blue}{0.660 / 0.493 / 88.8} \\
     & \cmark & Self-attention & \cmark & \textcolor{blue}{0.440 / 0.282 / 97.0} & 0.389 / 0.242 / 98.3 & \textcolor{red}{0.663 / 0.496 / 88.9} \\
    \midrule
    \multirow{4}{*}{\rotatebox[origin=c]{90}{TSCD}} & \xmark & - & \xmark & 0.278 / 0.163 / 89.7 & 0.295 / 0.173 / 93.7 & - \\
    & \cmark & Recurrent (RNN) & \xmark & 0.445 / 0.293 / 91.8 & 0.497 / 0.334 / 95.5 & - \\
    & \cmark & Recurrent (LSTM) & \xmark & \textcolor{red}{0.561 / 0.391 / 93.8} & \textcolor{blue}{0.565 / 0.394 / 96.6} & - \\
    & \cmark & Self-attention & \xmark & \textcolor{blue}{0.543 / 0.373 / 93.7} & \textcolor{red}{0.573 / 0.402 / 96.6} & - \\
    \bottomrule
  \end{tabular}
\end{table*}

\textbf{MTI module}: We perform a third ablation experiment investigating performance gains from leveraging additional edges in the MTI module. Here, we apply multi-task integration with different settings to the outputs of our method trained using the change loss with the maximum number of edges and the segmentation loss. The TFR module using self-attention was enabled. A degenerate Markov network that uses no change information (i.e., only the segmentation information represented as nodes) is added as a baseline. Table \ref{tab:ablation_mti} shows that introducing change information in the MTI module results in considerable change detection performance gains compared to the degenerate setting, as well as minor segmentation performance gains. For example, using the adjacent setting improves change detection performance (first-last) by 44.8 \% and 61.6 \% in terms of F1 score and IoU, respectively, compared to the degenerate setting on the SpaceNet 7 dataset. On the WUSU dataset, the corresponding performance improvements are 57.5 \% and 73.6 \% for the F1 score and IoU, respectively (cyclic loss scenario). Table \ref{tab:ablation_mti} also shows that introducing change information beyond adjacent edges results in further change detection performance gains, albeit to a much lesser extent. For example, compared to only using adjacent change information, dense information improved the F1 score and IoU values by 2.0 \% and 3.0 \%, respectively. In the context of the WUSU dataset, additional change information (i.e., adjacent vs. cyclic) improved the F1 score and IoU values by 1.6 \% and 2.2 \%, respectively. Therefore, the ablation study demonstrates that the proposed MTI module effectively integrates the outputs of the segmentation and change detection tasks at inference time.

\begin{table*}[!h]
\small
  \caption{Ablation results for the MTI module on the SpaceNet 7 (SN7) and WUSU datasets. The edge settings cyclic and dense are equivalent for the WUSU dataset since its time series consists of three images. The best and second-best performances are highlighted in red and blue, respectively.}
  \label{tab:ablation_mti}
  \centering
  \begin{tabular}{lclccc}
    \toprule
    \multirow{3}{*}{Dataset} & \multirow{3}{*}{MTI module} & \multirow{3}{*}{Edge setting} & \multicolumn{2}{c}{Change detection} & Segmentation \\
    & & & Bi-temporal & Continuous & \\
    & & & (F1 / IoU / OA) & (F1 / IoU / OA) & (F1 / IoU / OA) \\
    \midrule
    \multirow{5}{*}{\rotatebox[origin=c]{90}{SN7}} & \xmark & - & 0.537 / 0.367 / 98.8 & 0.397 / 0.248 / 99.7 & 0.593 / 0.422 / 94.3  \\
    & \cmark & degenerate & 0.373 / 0.229 / 97.8 & 0.173 / 0.095 / 98.8 & 0.593 / 0.422 / 94.3 \\
     & \cmark & adjacent & 0.540 / 0.370 / 99.0 & 0.410 / 0.258 / 99.7 & \textcolor{blue}{0.595 / 0.424 / 94.4}  \\
     & \cmark & cyclic & \textcolor{blue}{0.547 / 0.377 / 99.0} & \textcolor{blue}{0.412 / 0.260 / 99.7} & \textcolor{red}{0.596 / 0.424 / 94.3}  \\
     & \cmark & dense & \textcolor{red}{0.551 / 0.381 / 99.0} & \textcolor{red}{0.414 / 0.261 / 99.7} & \textcolor{red}{0.596 / 0.424 / 94.3}  \\
    \midrule
    \multirow{4}{*}{\rotatebox[origin=c]{90}{WUSU}} & \xmark & - & 0.420 / 0.266 / 96.6 & \textcolor{red}{0.391 / 0.243 / 98.3} & 0.660 / 0.493 / 88.8 \\
    & \cmark & degenerate & 0.275 / 0.159 / 93.8 & 0.203 / 0.113 / 95.5 & 0.660 / 0.493 / 88.8  \\
     & \cmark & adjacent & \textcolor{blue}{0.433 / 0.276 / 97.0} & \textcolor{blue}{0.389 / 0.242 / 98.3} & \textcolor{red}{0.663 / 0.497 / 88.9} \\
     & \cmark & cyclic & \textcolor{red}{0.440 / 0.282 / 97.0} & \textcolor{blue}{0.389 / 0.242 / 98.3} & \textcolor{blue}{0.663 / 0.496 / 88.9}  \\
    \bottomrule
  \end{tabular}
\end{table*}

\begin{figure}[!h]
     \centering
     \begin{subfigure}[b]{0.48\textwidth}
         \centering
         \includegraphics[width=\textwidth]{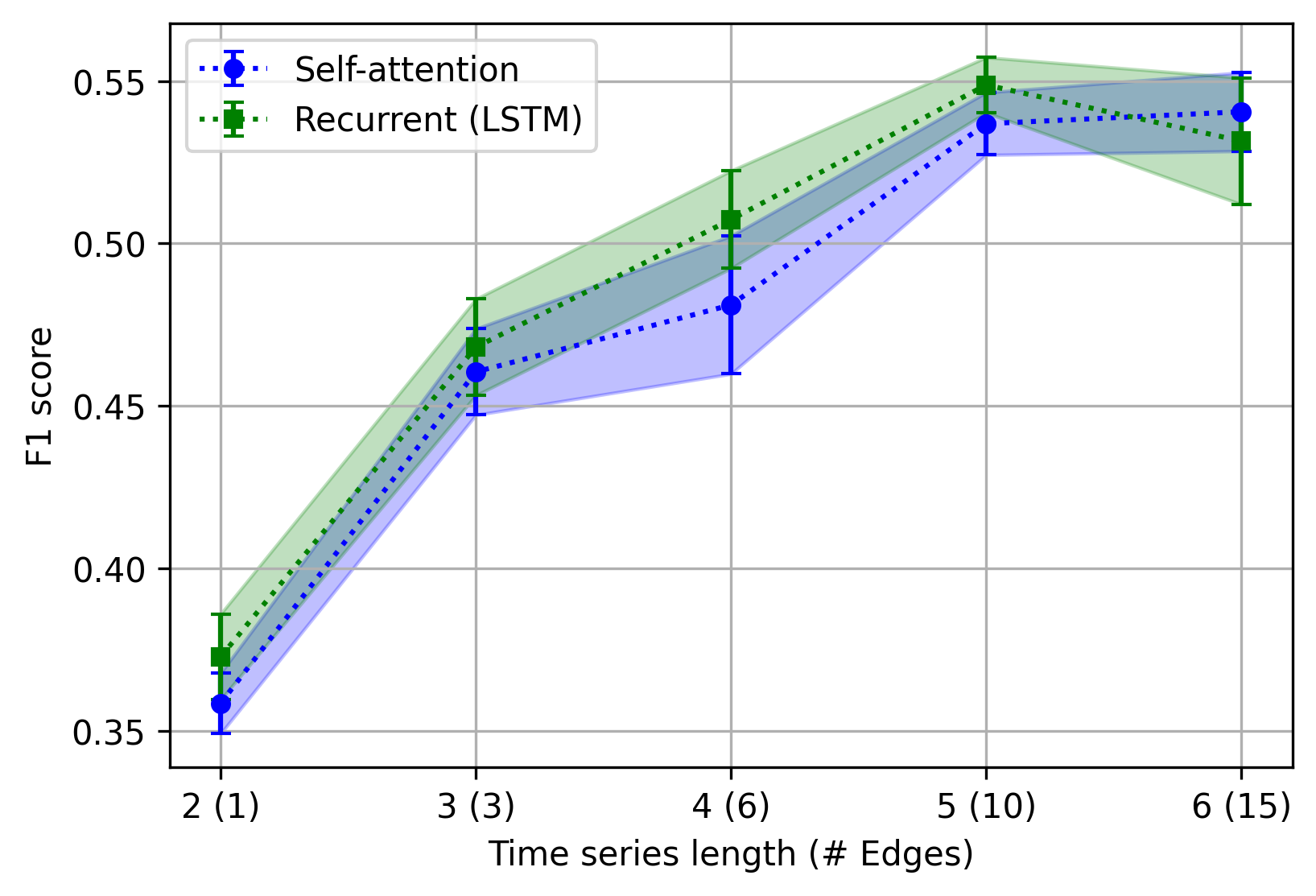}
         \caption{Bi-temporal change detection}
         \label{fig:ablation_3_flcd}
     \end{subfigure}
     \vspace{0.5cm}
     \vfill
     \begin{subfigure}[b]{0.48\textwidth}
         \centering
         \includegraphics[width=\textwidth]{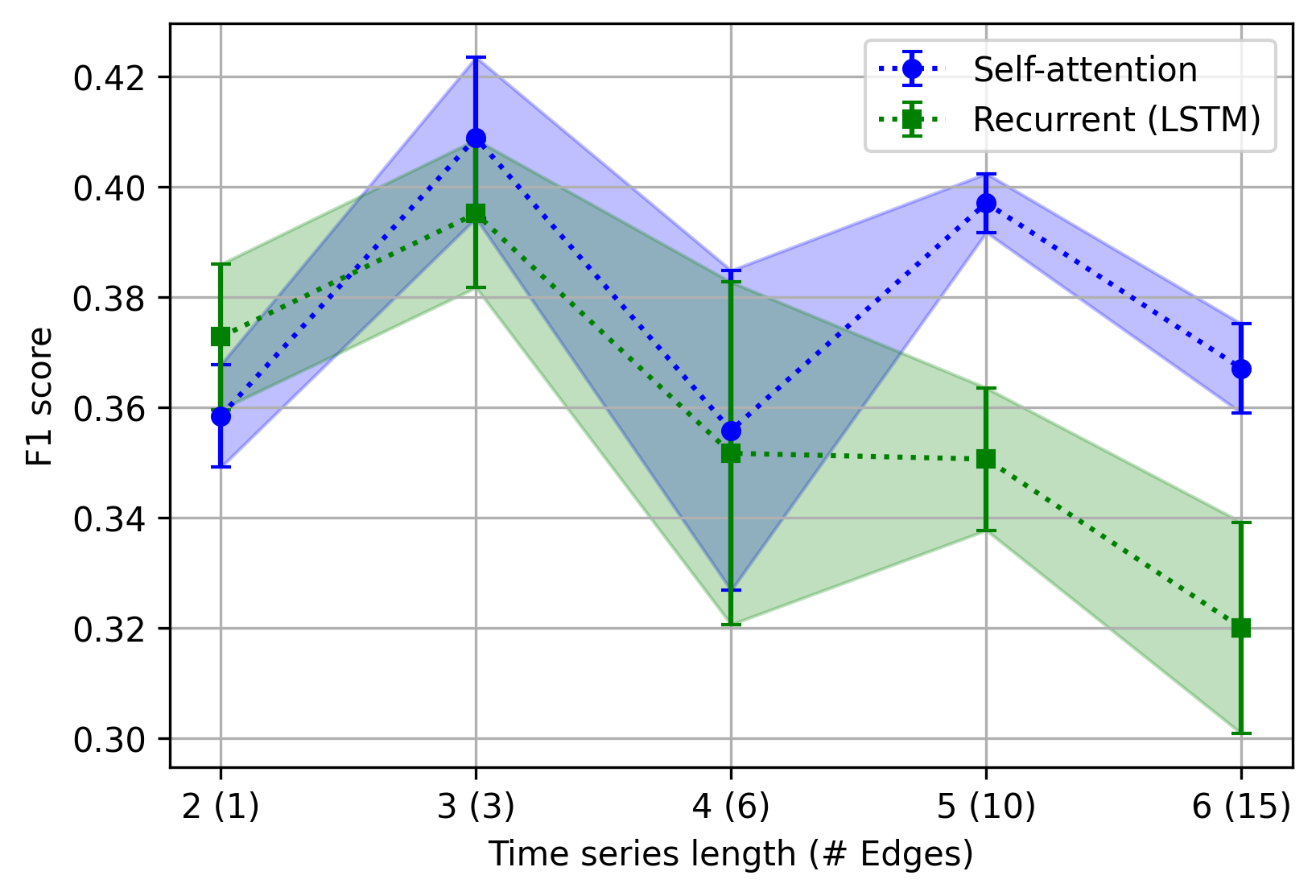}
         \caption{Continuous change detection}
         \label{fig:ablation_3_contcd}
     \end{subfigure}
     \vspace{0.5cm}
    \vfill
     \begin{subfigure}[b]{0.48\textwidth}
         \centering
         \includegraphics[width=\textwidth]{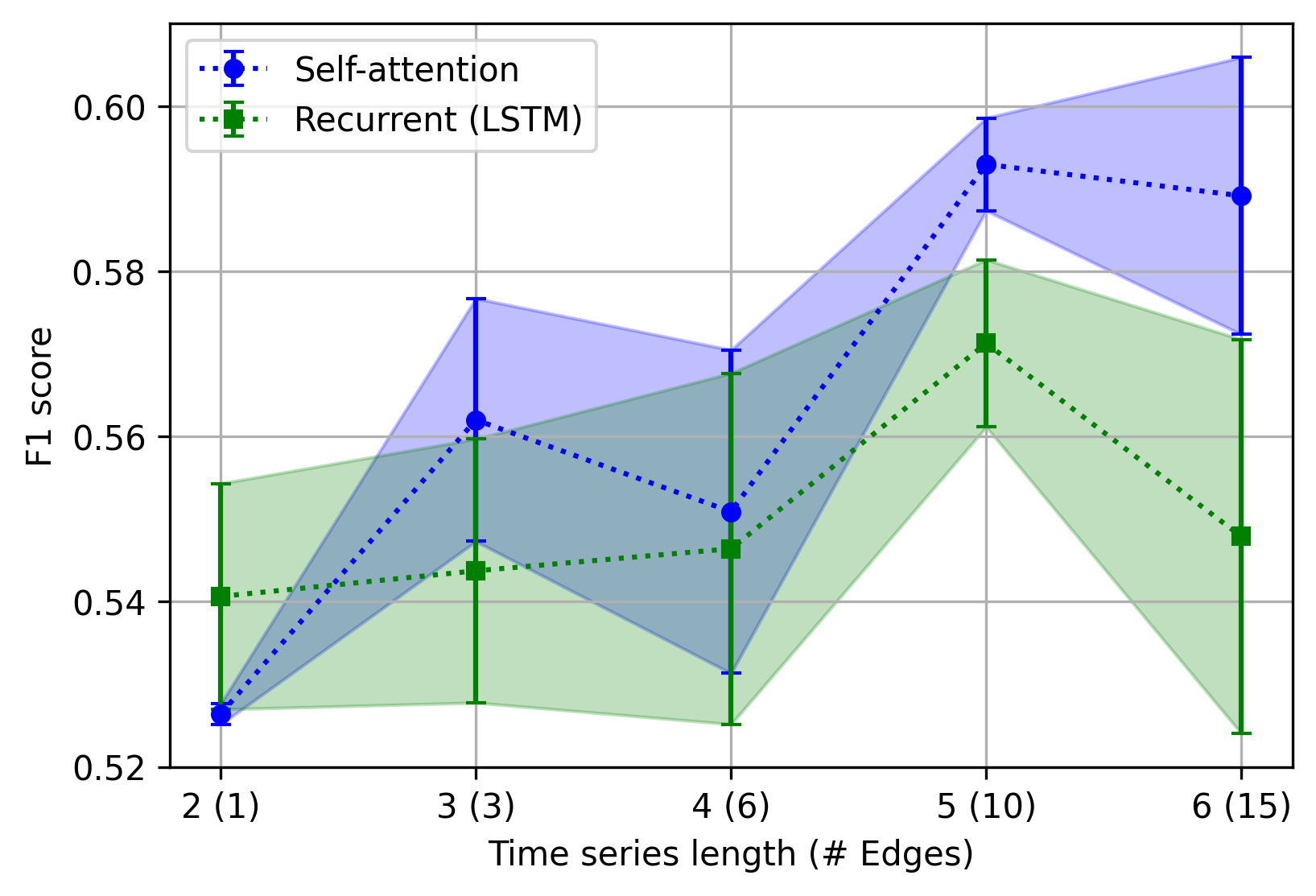}
         \caption{Segmentation}
         \label{fig:ablation_3_seg}
     \end{subfigure}
     \hfill
    \caption{The effect of time series length and sequence model in the TFR module on network performance (without MTI module) for the SpaceNet 7 dataset. Network performance is evaluated in terms of (a) bi-temporal change detection, (b) continuous change detection, and (c) segmentation. The maximum number of edges was used for each time series length in the change loss. Values represent the mean $\pm$ 1 standard deviation of 5 runs.}
    \label{fig:ablation_3}
\end{figure}

\textbf{Time series length}: To investigate the effect of SITS length on performance, we tested the proposed network with different settings for $T$ on SpaceNet 7. Additionally, we compared self-attention and LSTM for sequence modeling in the TFR module. The maximum number of edges (i.e., dense edge setting) was used across all $T$ settings in the change loss (see Equation \ref{eq:comb}). The segmentation loss was enabled, whereas the MTI module was disabled. It should be noted that for lengths $T=3$ and $T=2$, the edge settings dense are equivalent to cyclic and adjacent, respectively. Fig. \ref{fig:ablation_3} shows the results of this experiment for bi-temporal change detection (Fig. \ref{fig:ablation_3_flcd}), continuous change detection (Fig. \ref{fig:ablation_3_contcd}), and segmentation (Fig. \ref{fig:ablation_3_seg}). The bi-temporal results show that adding intermediate images improves change detection performance. However, performance saturates at $T=5$ and even decreases for the recurrent sequence model at $T=6$. Segmentation performance generally also increases with time series length until $T=5$. In contrast, the continuous change detection task increases in difficulty with time series length, since the temporal gap between adjacent image pairs in the time series decreases. Consequently, continuous change detection performance tends to decrease at longer time series lengths. Regarding the sequence model comparison, our network achieved better bi-temporal change detection performance using the recurrent model in the TFR module, except for the longest time series length ($T=6$). However, performance differences are below 5.5 \% across all time series lengths. On the other hand, self-attention outperformed the recurrent model for the continuous change detection and segmentation tasks. The largest performance gains in both cases were observed for the longest time series lengths (i.e., $T=5$ and $T=6$). 

\section{Discussion}
\label{sec:discussion}

Our results highlight the challenging nature of continuously detecting urban changes from high-resolution SITS. Indeed, most change detection studies focus on bi-temporal urban change detection from VHR imagery. The suite of methods considered SOTA for bi-temporal change detection typically employ the self-attention mechanism popularized by the transformer architecture to capture long-range contextual information in VHR imagery \cite{chen2020spatial, chen2021remote, bandara2022transformer, liu2023attention}. 
On popular benchmark datasets such as LEVIR-CD \cite{chen2020spatial} and WHU-CD \cite{ji2018fully}, these SOTA methods have achieved remarkable results with F1 scores exceeding 0.9. Although transformer-based methods generally outperformed their bi-temporal ConvNet-based counterparts in our experiments, they rarely achieved F1 scores higher than 0.4 on the continuous urban change detection task (Tables \ref{tab:quan_results_sn7}, \ref{tab:quan_results_wusu}, and \ref{tab:quan_results_tscd}). Therefore, our results indicate that bi-temporal change detection methods generally face limitations for continuous urban change detection.

In comparison to the SOTA methods for bi-temporal change detection, our method leverages the self-attention mechanism to model multi-temporal information in SITS. We showed that the proposed TFR module contributes to the network's representation learning capability, resulting in improved continuous change detection performance (Table \ref{tab:ablation_tfr}). We also compared our method against multi-temporal methods using self-attention or recurrent models for temporal modeling. However, these methods collapse the temporal dimension of the SITS, limiting them to the detection between the first and last image. Although multi-temporal methods generally outperformed bi-temporal methods on this task, they fell short of the proposed method (Tables \ref{tab:quan_results_sn7}, \ref{tab:quan_results_wusu}, and \ref{tab:quan_results_tscd}). Overall, we deem self-attention an effective mechanism for temporal modeling of SITS. Despite that, our ablation results show that recurrent sequence models such as LSTMs could be considered as an alternative, especially at shorter time series lengths (Table \ref{tab:ablation_tfr}).


Our work also highlights the need for effective integration approaches in multi-task learning schemes. Specifically, although multi-task learning is commonly applied for change detection \cite{liu2020building, papadomanolaki2021deep}, existing multi-task studies do not address the integration of the semantic segmentation and change detection outputs. To fill this research gap, we proposed the MTI module that represents segmentation and change predictions using Markov networks to find the optimal built-up area state for each timestamp in a pixel time series. Our results demonstrate that the proposed integration approach improves both tasks, namely the change detection and building segmentation task (Table \ref{tab:ablation_mti}). Furthermore, we demonstrate that the proposed approach benefits from integrating dense change information, obtained from predicting changes between all possible combinations of satellite image pairs in a time series, compared to using only adjacent change information (Table \ref{tab:ablation_mti}).  

Although we demonstrated the effectiveness of our method across three datasets, performances vary significantly across them. Overall, the lowest change detection performances were obtained for the WUSU dataset. The visualizations of the Gaofen-2 SITS highlight several challenging aspects for change detection in this dataset (Figures \ref{fig:qual_results_wusu_1} and \ref{fig:qual_results_wusu_2}). In particular, the images were acquired under different atmospheric conditions, and they have large off-nadir angles and contain shadows. In comparison, these artifacts are not apparent in the SpaceNet 7 images (Figures \ref{fig:qual_results_sn7_1} and \ref{fig:qual_results_sn7_2}), reducing the complexity of the change detection task. However, building segmentation performance on the SpaceNet 7 dataset is lower than on the WUSU dataset. We attribute this to the fact that individual buildings are more difficult to distinguish in PlanetScope imagery due to its lower spatial resolution (see Table \ref{tab:datasets}). The TSCD dataset stands out for its strong continuous urban change detection performances, while performances on the bi-temporal task are lower, especially for bi-temporal change detection methods. Here, it should be considered that the dataset only provides labels between adjacent images. Therefore, we had to derive change labels for non-adjacent images (see Section \ref{subsec:datasets}), which could affect the reference data quality.


Despite the improvements our method achieves over existing methods, we also identified several limitations related to our work. First of all, our integration approach relies on meaningful potentials extracted from the multi-task network outputs. However, the outputs of deep networks may not be well-calibrated \cite{guo2017calibration}. Furthermore, we assumed that our networks do not encounter out-of-distribution data during deployment due to the within-scene splits. In practice, however, urban mapping and change detection methods may encounter domain shifts when deployed to unseen geographic areas \cite{hafner2022unsupervised, hafner2023semi}. Therefore, future work will test the susceptibility of our multi-task integration approach to out-of-distribution data. For example, the effectiveness of the MTI module could be improved by explicitly calibrating the segmentation and change outputs of the model, using calibration techniques such as temperature scaling \cite{guo2017calibration}.

Another limitation of the proposed method is that it requires continuous building labels for training. Most popular urban change detection datasets are bi-temporal and feature VHR imagery. On the other hand, few urban change detection datasets featuring SITS and corresponding building labels for each image are available. Therefore, weakly supervised methods, using partial annotation or less accurate labeling, should be investigated for continuous urban change detection from SITS (e.g., \cite{meshkini2024multi}).

\section{Conclusion}
\label{sec:conclusion}

This study introduces a continuous urban change detection framework for optical SITS. The proposed method incorporates a transformer-based module to temporally refine feature representations extracted from image time series using a shared ConvNet. Unlike existing temporal modules in multi-temporal change detection methods, our module preserves the temporal dimension, enabling the detection of continuous changes. Additionally, we propose a novel multi-task integration (MTI) approach based on pairwise Markov networks, effectively combining building segmentation and dense urban change information. We evaluated our method on three SITS change detection datasets: SpaceNet 7, the WUSU dataset, and the TSCD dataset. The proposed method outperformed existing bi-temporal and multi-temporal change detection methods and segmentation methods. In particular, our findings show the limitations of bi-temporal methods in continuous change detection, as they cannot fully exploit multi-temporal information in SITS. While multi-temporal change detection methods overcome this limitation, they remain constrained to detecting changes between the first and last images in SITS. Our ablation study further demonstrates the effectiveness of the TFR module in modeling multi-temporal information and the benefits of incorporating dense change information during training. Moreover, it confirms that the MTI module successfully integrates segmentation and change outputs, leading to improved accuracy across both tasks. In summary, this research underscores the potential of high-resolution SITS for continuous urban change detection. Future work will explore weakly supervised and self-supervised change detection methods for SITS to reduce dependence on annotations.

\section*{Acknowledgement}
\label{sec:acknowledgement}
This work was supported by the EU Horizon 2020 project HARMONIA under agreement No. 101003517, Digital Futures under the grant for the EO-AI4GlobalChange project, and the EO-AI4Urban project within the ESA-MOST Dragon 5 Programme.

\bibliographystyle{IEEEtran}
\bibliography{refs}


 




\vfill

\end{document}